\documentclass[10pt,twocolumn,letterpaper]{article}

\usepackage[pagenumbers]{cvpr} 
\usepackage[dvipsnames]{xcolor}

\definecolor{cvprblue}{rgb}{0.21,0.49,0.74}
\usepackage[pagebackref,breaklinks,colorlinks,citecolor=cvprblue]{hyperref}

\usepackage{graphicx}
\usepackage{amsmath}
\usepackage{amssymb}
\usepackage{booktabs}
\usepackage{CJKutf8}  
\usepackage{array}
\usepackage{tabularx}
\usepackage{multirow}
\usepackage{subcaption}
\usepackage[dvipsnames]{xcolor}
\usepackage{breakcites} 
\usepackage{comment}
\usepackage{url}
\usepackage{colortbl}
\usepackage{nccmath}
\usepackage{makecell}
\usepackage{times}
\usepackage{epsfig}

\DeclareMathAlphabet\mathbfcal{OMS}{cmsy}{b}{n}

\usepackage[capitalize]{cleveref}
\crefname{section}{Sec.}{Secs.}
\Crefname{section}{Section}{Sections}
\Crefname{table}{Table}{Tables}

\newcolumntype{Y}{>{\centering\arraybackslash}X}

\def\confName{CVPR}
\def\confYear{2024}

\usepackage{xcolor}
\usepackage{wrapfig}
\usepackage{soul}

\definecolor{yellow}{rgb}{1,1, 0.6}
\definecolor{lightyellow}{rgb}{1,1, 0.8}
\definecolor{orange}{rgb}{1, 0.8, 0.6}
\definecolor{coral}{RGB}{246,131,65}
\definecolor{pinkred}{rgb}{1, 0.6, 0.6}
\definecolor{hotpink}{RGB}{238,64,195}
\definecolor{lavender}{RGB}{207,226,243}
\definecolor{gainsboro}{RGB}{208,224,227}
\definecolor{gainsboro2}{RGB}{217,234,211}
\definecolor{blanchedalmond}{RGB}{252,229,205}

\begin{document}
\CJKfamily{mj}

\title{Differentiable Display Photometric Stereo}

\author{Seokjun Choi\footnotemark[2] ~ ~ ~
Seungwoo Yoon\footnotemark[2] ~ ~ ~
Giljoo Nam\footnotemark[1] ~ ~ ~
Seungyong Lee\footnotemark[2] ~ ~ ~
Seung-Hwan Baek\footnotemark[2] \\
\footnotemark[2]~ POSTECH   ~ ~ ~   ~ ~ ~  \footnotemark[1]~ Meta Reality Labs\\
}

\twocolumn[{
\renewcommand\twocolumn[1][]{#1}
\maketitle
\vspace{-30pt}
\begin{center}
    \centering
    \captionsetup{type=figure}
    \includegraphics[width=\linewidth]{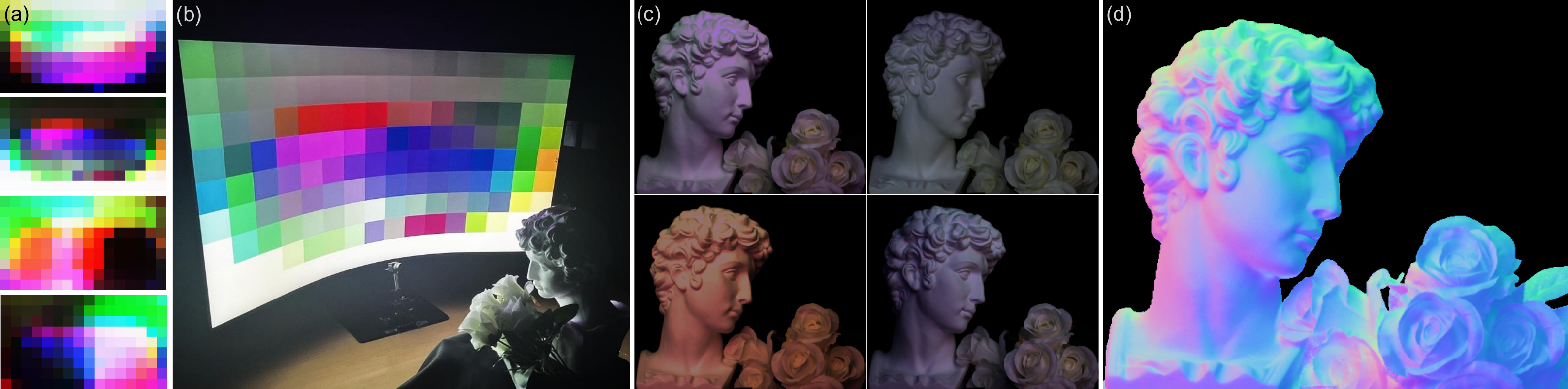}
    \vspace{-20pt}
    \captionof{figure}{We propose differentiable display photometric stereo, a method that facilitates (a) the learning of display patterns, enabling high-quality reconstruction of surface normals using (b) a monitor and a camera. (c) Capturing a scene with the learned patterns allows for estimating (d) high-quality surface normals.
}
    \label{fig:teaser}
    \vspace{-3mm}
\end{center}
}]

\begin{abstract}
\vspace{-5mm}
Photometric stereo leverages variations in illumination conditions to reconstruct surface normals.
Display photometric stereo, which employs a conventional monitor as an illumination source, has the potential to overcome limitations often encountered in bulky and difficult-to-use conventional setups. 
In this paper, we present differentiable display photometric stereo (DDPS), addressing an often overlooked challenge in display photometric stereo: the design of display patterns. 
Departing from using heuristic display patterns, DDPS learns the display patterns that yield accurate normal reconstruction for a target system in an end-to-end manner.
To this end, we propose a differentiable framework that couples basis-illumination image formation with analytic photometric-stereo reconstruction. 
The differentiable framework facilitates the effective learning of display patterns via auto-differentiation.
Also, for training supervision, we propose to use 3D printing for creating a real-world training dataset, enabling accurate reconstruction on the target real-world setup.
Finally, we exploit that conventional LCD monitors emit polarized light, which allows for the optical separation of diffuse and specular reflections when combined with a polarization camera, leading to accurate normal reconstruction.
Extensive evaluation of DDPS shows improved normal-reconstruction accuracy compared to heuristic patterns and demonstrates compelling properties such as robustness to pattern initialization, calibration errors, and simplifications in image formation and reconstruction.
\end{abstract}

\section{Introduction}
\label{sec:intro}
Reconstructing high-quality surface normals is pivotal in computer vision and graphics for 3D reconstruction~\cite{ma2007rapid,park2016robust}, relighting~\cite{meka2020deep,pandey2021total}, and inverse rendering~\cite{schmitt2020joint,zhang2022iron}. Among various techniques, photometric stereo~\cite{woodham1980photometric} leverages the intensity variation of a scene point under varied illumination conditions to reconstruct normals. Photometric stereo finds its application in various imaging systems including light stages~\cite{weyrich2006analysis,legendre2016practical,meka2019deep,zhou2023relightable}, handheld-flash cameras~\cite{nam2018practical,zhang2022iron,azinovic2022polface,cheng2023wildlight}, and display-camera systems~\cite{aittala2013practical,sengupta2021light,lattas2022practical}.

Display photometric stereo uses monitors and cameras as a versatile and accessible system that can be conveniently placed on a desk~\cite{aittala2013practical,sengupta2021light,lattas2022practical}. 
Producing diverse illumination conditions can be simply achieved by displaying multiple patterns using pixels on the display as programmable point light sources.
This convenient and intricate modulation of illumination conditions significantly enlarges the design space of illumination patterns for display photometric stereo. 
Nevertheless, existing approaches often rely on heuristic display patterns, resulting in sub-optimal reconstruction quality. 

In this paper, to exploit the large design space of illumination patterns in display photometric stereo, we propose differentiable display photometric stereo (DDPS).
The key idea is to learn display patterns that lead to improved reconstruction of surface normals for a target system in an end-to-end manner.
To this end, we introduce a differentiable framework that combines basis-illumination image formation and an optimization-based photometric stereo method. 
This enables effective pattern learning by directly optimizing the display patterns via auto-differentiation.
To compute the normal-reconstruction loss for backpropagation, we propose the use of 3D printing for creating a real-world training dataset with known geometry. 
Combined with the basis-illumination image formation, using the 3D-printed dataset allows for efficient and realistic simulation of relit images during end-to-end optimization.
In addition, we leverage that conventional LCD monitors emit polarized light. Thus, using a polarization camera, we can optically remove specular reflection that often deteriorates photometric-stereo reconstruction.

Extensive evaluation of DDPS on diverse objects shows that using the learned patterns significantly improves normal accuracy compared to using heuristic patterns. 
Moreover, DDPS exhibits robustness to pattern initialization, calibration error, and simplifications in image formation and reconstruction, promising {its} practical applicability. 
We will release code and data upon acceptance.

In summary, our contributions are as follows:
\begin{itemize}
\item Departing from using heuristic patterns for display photometric stereo, we directly learn display patterns that lead to high-quality normal reconstruction for display photometric stereo in an end-to-end manner. 
\item For DDPS, we propose the differentiable framework {consisting} of basis-illumination image formation and analytic photometric-stereo reconstruction, the use of 3D-printed objects for a training dataset, and using a polarized LCD and a polarization camera.

\item We perform extensive experiments, demonstrating the effectiveness of learned patterns, which outperforms heuristic patterns, and the robustness of DDPS against various factors including pattern initialization and calibration errors. 
\end{itemize}

\section{Related Work}
\label{sec:related}

\begin{figure*}[ht]
	\centering
		\includegraphics[width=0.9\linewidth]{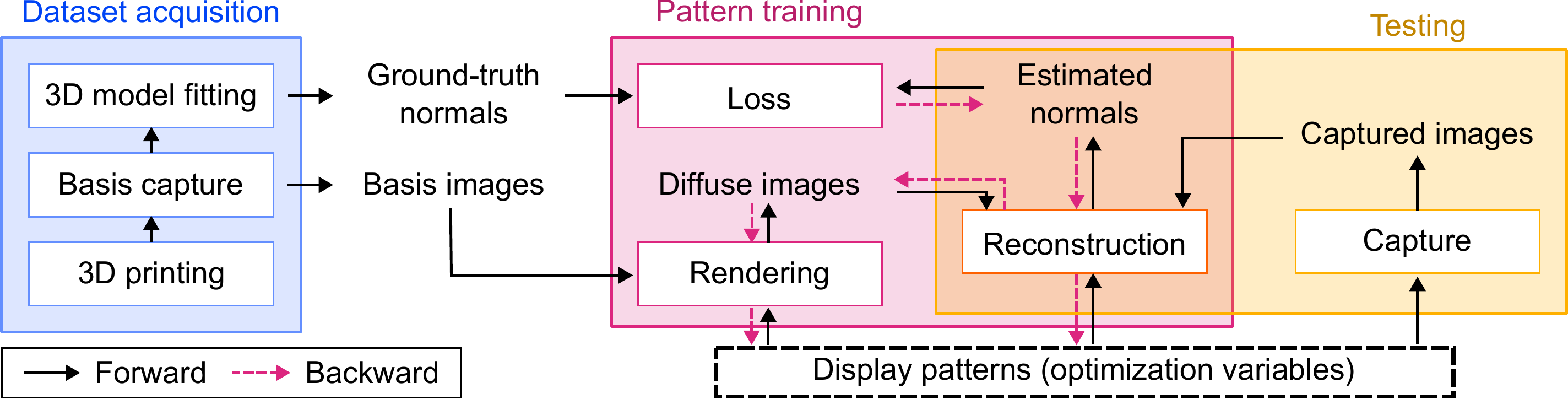}
		\caption{\textbf{Overview of DDPS.} DDPS consists of three stages: dataset acquisition, pattern training, and testing.}
  \vspace{-4mm}
  		\label{fig:overview}
\end{figure*}

\paragraph{Illumination Patterns for Photometric Stereo}
One crucial but often overlooked problem in photometric stereo is deciding on illumination patterns, which is a set of intensity distributions of light sources, so that accurate surface normals can be reconstructed.
A standard option is the one-light-at-a-time (OLAT) pattern that turns on each light source at its maximum intensity one by one~\cite{sun2020light,zhang2021neural}. OLAT is typically employed when the intensity of each light source is sufficient enough to provide light energy to be detected by a camera sensor without significant noise, such as in light stages~\cite{debevec2000acquiring}.
Extending OLAT patterns with a group of neighboring light sources increases light energy, reducing measurement noise~\cite{wenger2005performance,bi2021deep}.
Spherical gradient illumination, designed for light stages, enables rapid acquisition of high-fidelity normals by exploiting polarization~\cite{ma2007rapid}, color~\cite{meka2019deep}, or both~\cite{fyffe2015single}.
Complementary patterns, where half of the lights are turned on and the other half off for each three-dimensional axis, also enable rapid reconstruction when applied to light stages and monitors~\cite{kampouris2018diffuse,lattas2022practical}.
Wenger et al.\cite{wenger2005performance} propose random binary patterns that provide high light efficiency.
However, the aforementioned illumination patterns are heuristically designed, which often result in sub-optimal reconstruction accuracy and capture efficiency.
For a specific display-camera system, it is challenging to determine which display patterns would provide high-quality photometric stereo.
DDPS departs from using heuristic patterns and instead learns display patterns for robust photometric stereo.

\paragraph{Illumination-optimized Systems}
Recent studies have investigated optimizing illumination designs for inverse rendering~\cite{kang2018efficient,kang2019learning,ma2021free,zhang2023deep}, active-stereo depth imaging~\cite{baek2021polka}, and holographic display~\cite{peng2020neural}.
These approaches typically rely on dedicated illumination modules such as LED arrays, diffractive optical elements, and spatial light modulators.
In contrast, DDPS exploits ubiquitous LCD devices and their polarization state for display illumination.
Also, DDPS directly applies normal reconstruction loss to illumination learning using the 3D-printed dataset, unlike previous method that employ intermediary metrics, such as lumitexel prediction~\cite{kang2018efficient,kang2019learning,ma2021free}. 
Zhang et al.\cite{zhang2023deep} optimize a single illumination pattern for inverse rendering, only targeting planar objects.
In contrast, DDPS reconstructs surface normals of general objects with complex shapes and capable of optimizing multiple illumination patterns. 

\paragraph{Imaging Systems for Photometric Stereo}
Many photometric stereo systems have been proposed, including moving a point light source, such as a flashlight on a mobile phone~\cite{riviere2016mobile,hui2017reflectance}, a DSLR camera flash~\cite{fyffe2016near,deschaintre2021deep}. and installing multiple point light sources in light stage systems~\cite{legendre2016practical,meka2019deep} and other custom devices~\cite{havran2017lightdrum,kang2018efficient,kang2019learning,ma2021free, kampouris2018diffuse}. 
Display photometric stereo exploits off-the-shelf displays as cost-effective, versatile active-illumination modules capable of generating spatially-varying trichromatic intensity variation~\cite{aittala2013practical,ghosh2009estimating,clark2010photometric,francken2008high,lattas2022practical,liu2018near,nogue2022polarization}. Lattas et al.~\cite{lattas2022practical} demonstrated facial capture using multiple off-the-shelf monitors and multi-view cameras with trichromatic complementary illumination.
In our paper, we build on display photometric stereo and propose to learn the display patterns to obtain high-quality normal reconstruction.

\paragraph{Photometric Stereo Dataset}
Many datasets have been proposed for photometric stereo~\cite{ren2022diligent102,li2020multi,mecca2021luces,xiong2014shading,alldrin2008photometric}  for evaluation or training photometric stereo methods.
Early datasets often relied on synthetic rendering~\cite{santo2017deep,chen2020learned}.
However, using synthetic datasets for a real-world target system requires highly accurate calibration of the target photometric-stereo system, its replication on the rendering, and physically realistic light-transport simulation. 
Real-world datasets relax these constraints by capturing real-world objects under multiple point light sources~\cite{ren2022diligent102,li2020multi}.  
However, acquiring ground-truth normals of real-world objects often demands using high-quality commercial 3D scanners. 
In contrast, DDPS uses 3D printing to obtain objects with known geometry. Using the 3D-printed dataset combined with {3D model fitting} allows for effectively supervising the pattern learning in an end-to-end manner. 
\section{Overview}
\label{sec:overview}
DDPS consists of three stages as shown in Figure~\ref{fig:overview}: dataset acquisition, pattern training, and testing.
First, in the dataset-acquisition stage, we 3D-print various objects, capture their basis-illumination images with a target display-camera setup, and obtain ground-truth surface normal maps via {3D model fitting.}
Then, in the pattern-training stage, we learn the {optimal} display patterns that lead to high-quality normal reconstruction using the real-world training dataset. To this end, we develop the differentiable framework of basis-illumination image formation and analytic photometric-stereo reconstructor. 
In the testing phase, we capture diverse real-world objects under the patterns learned on our training dataset and reconstruct surface normals using the photometric-stereo reconstructor. 
\section{Polarimetric Display-Camera Imaging}
\label{sec:system}
\begin{figure}[t]
    \includegraphics[width=\linewidth]{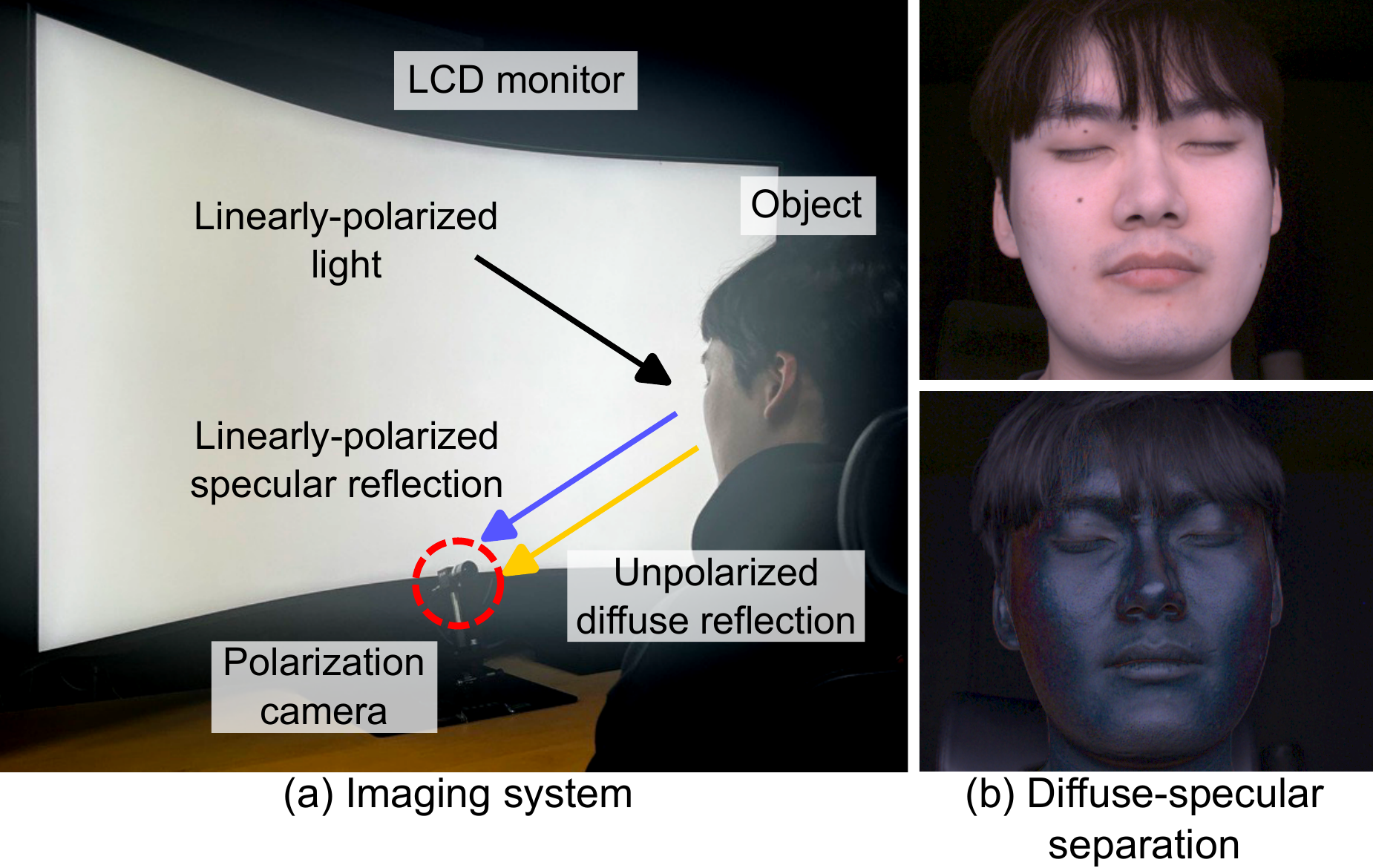}
    \vspace{-5mm}
    \caption{\textbf{Polarimetric imaging system.} (a) Imaging system consisting of an LCD monitor and a polarization camera. Decomposed (b) diffuse image and specular image using linearly-polarized light emitted from the monitor.}
    \vspace{-4mm}
    \label{fig:system}
\end{figure}

\paragraph{Polarimetric Light Transport}
We first describe our imaging system, shown in Figure~\ref{fig:system}(a). 
We use off-the-shelf components: a curved {4K} LCD monitor and a polarization camera. 
Linearly-polarized light is emitted from the LCD monitor, due to the polarization-based working principle of LCDs~\cite{collett2005field}. The light interacts with a real-world scene, generating both specular and diffuse reflections. The specular reflection tends to maintain the polarization state of light, while diffuse reflection becomes unpolarized~\cite{baek2020image}.
The polarization camera then captures the reflected light at four different linear-polarization angles: $\{I_\theta\}_{\theta \in \{0^\circ, 45^\circ, 90^\circ, 135^\circ\}}$.
We then convert the captured raw intensities $\{I_\theta\}$ into the linear Stokes-vector elements~\cite{collett2005field}: 
\begin{align}
\label{eq:polar_decomp}
&s_0 = \frac{\sum_\theta{I_{\theta}}}{2}, \, s_1 = I_{0^\circ} - I_{90^\circ}, \, s_2 = 2  I_{45^\circ} - I_{0^\circ}, 
\end{align}
and compute the diffuse reflection $I_\text{diffuse}$ and specular reflection $I_\text{specular}$:
$I_\text{specular} = \sqrt{s_1^{2} + s_2^{2}},\quad I_\text{diffuse} = s_0 - I_\text{specular}$.
Hereafter, we will denote $I \leftarrow I_\text{diffuse}$ as the diffuse image obtained by the polarimetric decomposition.
Figure~\ref{fig:system}(b) shows the separated diffuse and specular images. The diffuse image $I$ will be used for photometric stereo. 
Note that this diffuse-specular separation using polarized illumination and cameras has been often used in other systems~\cite{francken2008high,ghosh2009estimating} such as light stages. DDPS applies the same principle to the display photometric stereo by using a conventional LCD and a polarization camera. 

\paragraph{Display Superpixels}
For the computational efficiency of our end-to-end optimization, we parameterize the display with $P=16\times9$ superpixels, where each superpixel is a group of $240 \times 240$ pixels. 
Ablation on the superpixel resolution can be found in the Supplemental Document.

\paragraph{Calibration}
We estimate the location of each superpixel with respect to the camera.
To this end, we develop a mirror-based calibration method that estimates superpixel locations by using display patterns reflected on a mirror.
We refer to the Supplemental Document for the details on the mirror-based calibration.
We also calibrate the intrinsic parameters of the camera and the non-linearity of display intensity using standard methods~\cite{zhang2000flexible}.
Figure~\ref{fig:superpixel_cal} shows the calibrated superpixel locations.
\section{Dataset Creation using 3D Printing}
\label{sec:dataset}
\begin{figure}[t]
\centering
\includegraphics[width=\linewidth]{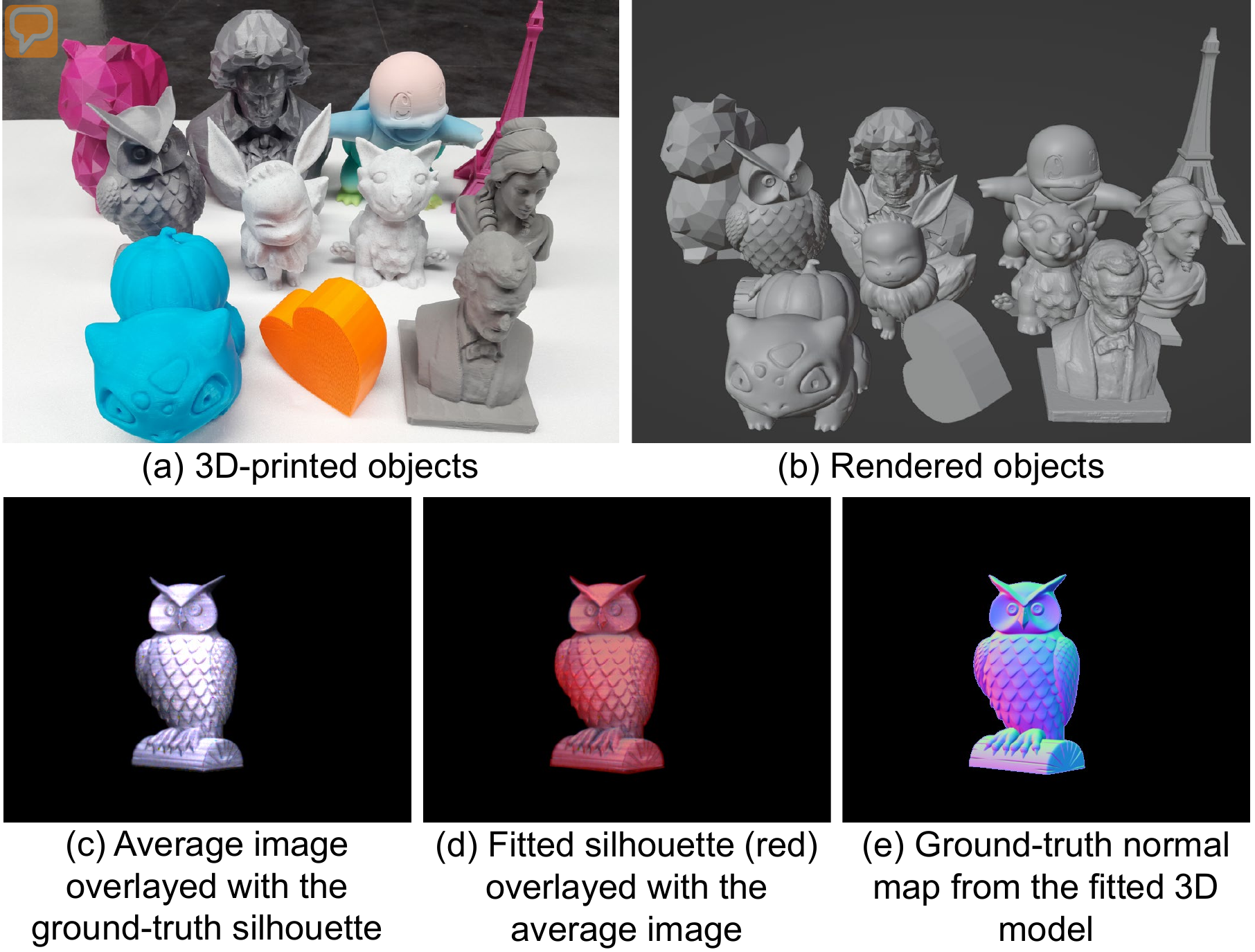}
\vspace{-5mm}
\caption{
\textbf{Training dataset creation with 3D printing.} To learn display patterns, we propose to use (a) 3D-printed objects that have corresponding (b) known ground-truth 3D models.
(c) We extract the silhouette $S$ from the averaged basis images and (d) align the ground-truth 3D models with the captured image as depicted with the fitted silhouette in red on top of the average image. 
(e) We obtain a ground-truth normal map from the fitted 3D model.
}
\vspace{-4mm}
\label{fig:dataset}
\end{figure}

We describe our strategy for creating a training dataset using 3D printing.
This allows for easily creating a real-world dataset with known geometry that can be used for DDPS. 
Figure~\ref{fig:dataset}(a)\&(b) show the 3D printed objects and their ground-truth 3D models.
For each training scene, we capture {raw} basis images $\mathbfcal{B}=\{B_j\}_{j=1}^P$, where $j$ is the index of the basis illumination of which {only} $j$-th superpixel is turned on with its full intensity as white color. 
We then extract the silhouette mask $S$ using the average image of the basis images $I_\mathrm{avg}$ that present well-lit appearance for most of the object scene points as shown in Figure~\ref{fig:dataset}(c). 
Given the silhouette mask $S$, we align the ground-truth geometry of the 3D-printed object in the scene by optimizing the pose of the ground-truth mesh with a silhouette rendering loss:

\begin{equation}
\label{eq:opt_dataset}
\underset{\mathbf{t}, \mathbf{r}}{\text{minimize}} \| f_{s}(\pi; \mathbf{t}, \mathbf{r}) - S \|_2^2,
\end{equation}
where $\pi$ is the known 3D model, $\mathbf{t}$ and $\mathbf{r}$ are the translation and rotation of the model. $f_{s}(\cdot)$ is the differentiable silhouette rendering function. 
We solve Equation~\eqref{eq:opt_dataset} using gradient descent in Mitsuba3~\cite{jakob2022mitsuba3}. 
Once the pose parameters are obtained, we render the normal map with the 3D model at the optimized pose, which serves as the ground-truth normal map $N_\mathrm{GT}$, shown in Figure~\ref{fig:dataset}. 
We create 40 training {scenes} and 4 test {scenes} with ground-truth normals. 
Note that even trained on the 3D-printed objects, DDPS enables effective reconstruction for diverse real-world objects as demonstrated in the results. 

\begin{figure*}[t]
    \centering
    \includegraphics[width=\linewidth]{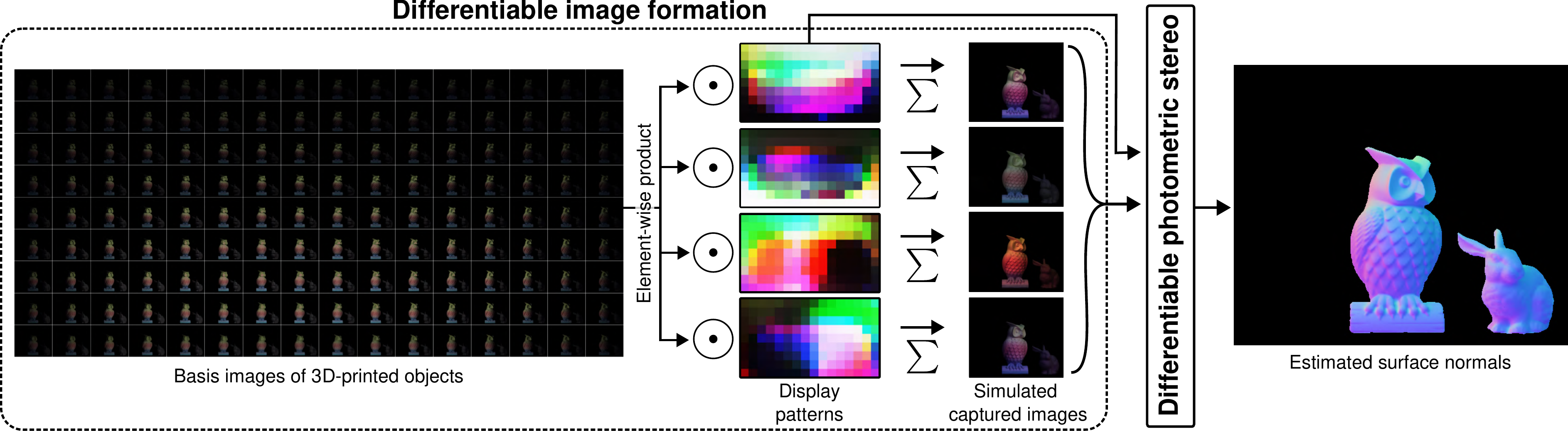}
    \caption{\textbf{Differentiable framework.} Using 3D-printed objects as a dataset allows for simulating real-world captured images with a differentiable image formation. We reconstruct high-fidelity surface normals using differentiable photometric stereo from the simulated captured images.} 
    \label{fig:learning} 
\vspace{-3mm}
\end{figure*}

\section{Learning Display Patterns}
\label{sec:e2e}
We learn display patterns using the 3D-printed training dataset consisting of ground-truth normal maps $N_\mathrm{GT}$ and basis images $\mathbfcal{B} = \{B_j\}_{j=1}^P$. 
We denote $K$ different display patterns as $\mathbfcal{M}=\{\mathcal{M}_i\}_{i=1}^K$, where the $i$-th display pattern $\mathcal{M}_i$ is modeled as an RGB intensity pattern of $P$ superpixels: $\mathcal{M}_i \in \mathbb{R}^{P\times3}$, which is our optimization variable. 

For end-to-end training of the display RGB intensity patterns $\mathbfcal{M}$, we develop a differentiable image formation function $f_I(\cdot)$ and a differentiable photometric-stereo method $f_n(\cdot)$, which are chained together via auto-differentiation. The differentiable image formation $f_I(\cdot)$ takes a display pattern $\mathcal{M}_i$ and the basis images $\mathbfcal{B}$ of a training scene, and simulates the captured images $\mathbfcal{I}=\{I_i\}_{i=1}^K$ for the display patterns being optimized. 
The photometric-stereo method $f_n(\cdot)$ then processes the simulated captured images $\mathbfcal{I}$ to estimate surface normal $N$. 
Below, we describe each component in details.

\subsection{Differentiable Image Formation}
\label{sec:image_formation}
For the basis images $\mathbfcal{B}$ of a training sample, we simulate a raw image captured under a display pattern $\mathcal{M}_i$ as
\begin{align}
\label{eq:image_formation}
I_i = f_I(\mathcal{M}_i, \mathbfcal{B}) = \sum_{j=1}^{P} B_j \mathcal{M}_{i,j},
\end{align}
where $\mathcal{M}_{i,j}$ is the $j$-th superpixel RGB intensity in the display pattern $\mathcal{M}_i$.
For $K$ display patterns, we synthesize each image as 
\begin{align}
\label{eq:image_formation_all}
\mathbfcal{I} = \{f_I(\mathcal{M}_i,\mathbfcal{B})\}_{i=1}^K.
\end{align}
Figure~\ref{fig:learning} shows the overview of our image formation.

This weighted-sum formulation exploits the basis images acquired for real-world 3D printed objects, based on light-transport linearity in the regime of ray optics. Compared to using variants of rendering equations as differentiable image formations~\cite{baek2021polka,baek2022all}, {the image formation with basis images synthesizes realistic images in a computationally efficient manner, comprising only a single weighted summation, \emph{serving as {a memory-efficient and realistic image formation suitable for end-to-end pattern learning}.}

\subsection{Differentiable Photometric Stereo}
We reconstruct surface normal $N$ from the images $\mathbfcal{I}$ captured or simulated under the display patterns $\mathbfcal{M}$:
\begin{equation}
\label{eq:recon}
N=f_n(\mathbfcal{I},\mathbfcal{M}).
\end{equation}

Note that the images $\mathbfcal{I}$ mostly contain diffuse-reflection components as a result of the polarimetric diffuse-specular separation described in Section~\ref{sec:system}.
Using the optically-separated diffuse image $\mathbfcal{I}$, we develop an {analytic} trinocular photometric-stereo method that is {independent of the training dataset and has no training parameters}.
This enables \emph{effective end-to-end learning of display patterns by sorely focusing on optimizing display patterns without any other learning variables such as neural networks.}

We start by denoting the captured diffuse RGB intensity of a camera pixel under the $i$-th display pattern as $I_{i}^c$, where $c$ is the color channel $c \in \{R,G,B\}$.
Note that dependency on the pixel is omitted in the notation of $I_{i}^c$ for simplicity.
We denote the spatially-varying per-pixel illumination vector coming from the center of $j$-th superpixel on the monitor to a scene point corresponding to the pixel as $l_j$.
Note that the illumination vectors are computed considering the different locations of the scene points. The scene points are assumed to lie on a plane which is a fixed distance (50\,cm in our experiment) away from the camera. 
We then formulate a linear equation as
\begin{align}
\label{eq:photometric_stereo_rgb_matrix_normal}
\mathbf{I} = \boldsymbol{\rho}\odot\mathbf{M}\mathbf{l}\mathbf{N},
\end{align}
where {$\mathbf{I} \in \mathbb{R}^{3K\times1}$, $\boldsymbol{\rho} \in \mathbb{R}^{3K\times1}$, and $\mathbf{N} \in \mathbb{R}^{3\times1}$} are the vectorized intensity, albedo, and surface normals.
$\odot$ is Hadarmard product.
$\mathbf{M} \in \mathbb{R}^{3K\times P}$, $\mathbf{l} \in \mathbb{R}^{P\times3}$} are the matrices for the pattern intensity and illumination directions.
Note that the only unknown variables are the surface normal $\mathbf{N}$ and the albedo $\boldsymbol{\rho}$.
Refer to the Supplemental Document for the formulation details.

We set the albedo $\boldsymbol{\rho}$ as the max intensities among captures to for numerical stability and solve for the surface normal $\mathbf{N}$ using the pseudo-inverse method: $\mathbf{N} \leftarrow (\boldsymbol{\rho} \odot \mathbf{M} \mathbf{l})^{\dag}\mathbf{I}$, where $\dag$ is the pseudo-inverse operator.
Figure~\ref{fig:learning} shows the reconstructed surface normals. 
We exploit the differentiability of our analytic reconstructor for effective end-to-end optimization of display patterns. 

\subsection{Training}
Equipped with the image formation and the reconstructor, we learn the display patterns $\mathbfcal{M}$ by solving an optimization problem: 
\begin{equation}
\label{eq:opt_e2e}
\underset{\mathbfcal{M}}{\text{minimize}} \sum_{\mathbfcal{B}, N_\mathrm{GT}} \texttt{loss}\left( f_n\left( \{f_{I} (\mathcal{M}_i,\mathbfcal{B})\}_{i=1}^K, \mathbfcal{M} \right), N_\mathrm{GT} \right),
\end{equation}
where $\texttt{loss}(\cdot)=(1-N \cdot N_\mathrm{GT})/2$, which is the normalized cosine distance, penalizes the angular difference between the estimated and the ground-truth normals from the 3D-printed dataset, meaning that the patterns are learned on the entire training dataset. 
To ensure the physically-valid intensity range from zero to one of the display pattern $\mathbfcal{M}$, we apply a sigmoid function to the optimization variable: $\mathbfcal{M} \leftarrow \texttt{sigmoid}(\mathbfcal{M})$.
We use Adam optimizer~\cite{adam2015}.

\subsection{Testing}
Once the display patterns are learned, we perform testing on real-world objects. 
Specifically, we capture images under the learned $K$ display patterns, perform diffuse-specular separation, and obtain diffuse image $I_i$ for the $i$-th display pattern. 
We then estimate surface normals using our photometric stereo method: 
\begin{equation}
N = f_n(\mathbfcal{I}).
\end{equation}

\vspace{-6mm}
\section{Assessments}
\label{sec:assessment}

We assess DDPS on diverse objects. Refer to the Supplemental Document for {complete} results.

\begin{figure}[t]
    \centering
    \includegraphics[width=0.99\linewidth]{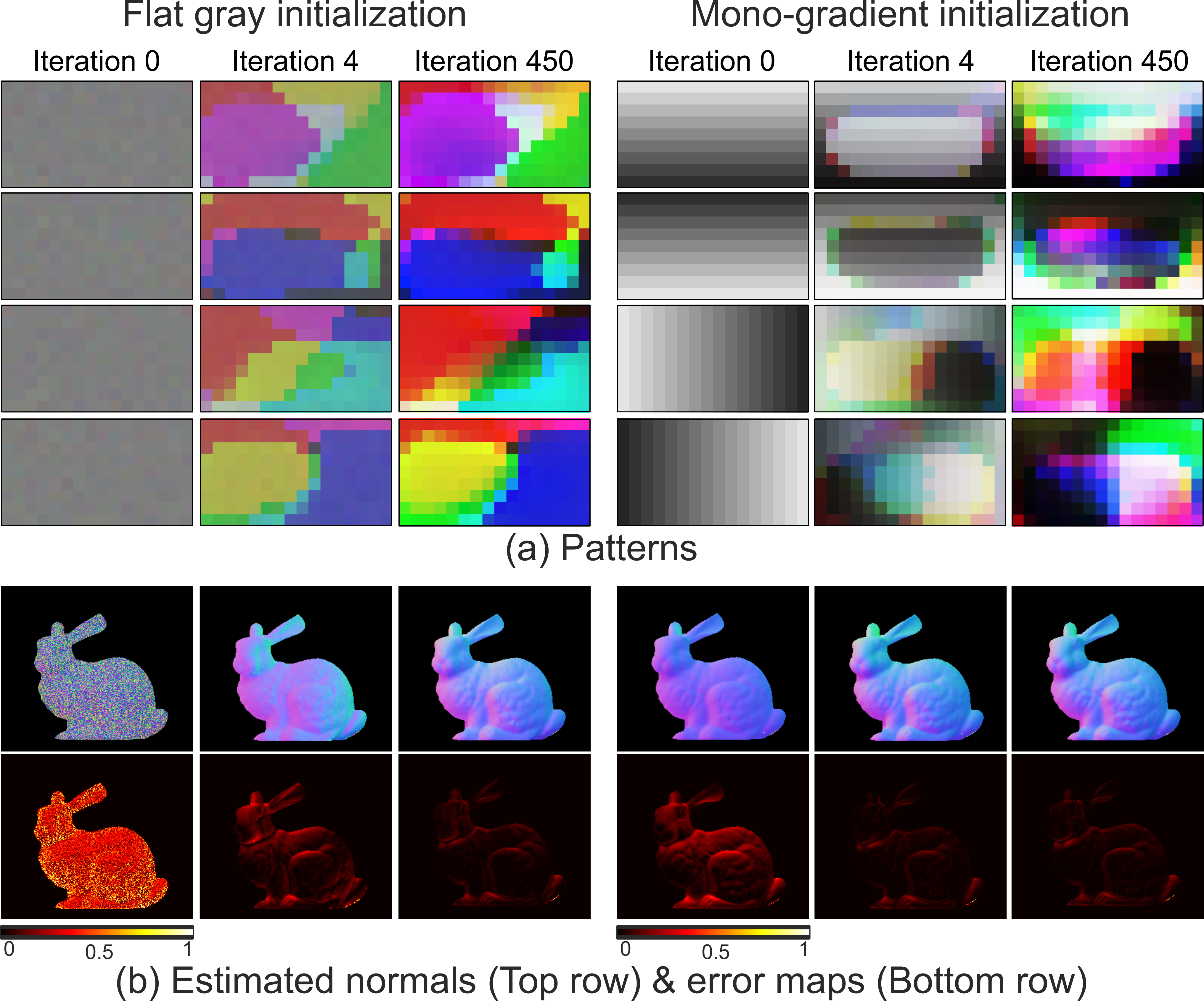}
    \caption{
    {
    \textbf{Learning process.} DDPS allows the learning of display patterns for high-quality normal reconstruction, not only from sub-optimal heuristically-designed patterns but also from flat gray noise that does not require any prior knowledge of the imaging system. 
    }
    }
    \vspace{-4mm}
    \label{fig:learned_patterns}
\end{figure}

\paragraph{Learned Patterns}
Figure~\ref{fig:learned_patterns} shows the patterns learned with DDPS. 
The learned patterns exhibit distinctively-colored regions and adjusted brightness for robust normal reconstruction.
We evaluate the learned patterns regarding normal-reconstruction accuracy with common heuristic patterns: OLAT~\cite{sun2020light}, group OLAT~\cite{bi2021deep}, monochromatic gradient~\cite{ma2007rapid}, monochromatic complementary~\cite{kampouris2018diffuse}, trichromatic gradient~\cite{meka2019deep}, trichromatic complementary~\cite{lattas2022practical}. 
Table~\ref{tab:learned_illum} show that the learned patterns outperform all existing heuristic patterns {on the test dataset}.
We measured the average reconstruction error $\texttt{loss}(\cdot)$ on the 3D-printed test dataset.
It is worth noting that \emph{DDPS allows using initial patterns that do not require any prior knowledge of the imaging system}.
That is, initialization with monochromatic random, trichromatic random, and flat gray noise also results in competitive results.

\begin{figure*}[!t]
 	\centering
 		\includegraphics[width=\linewidth]{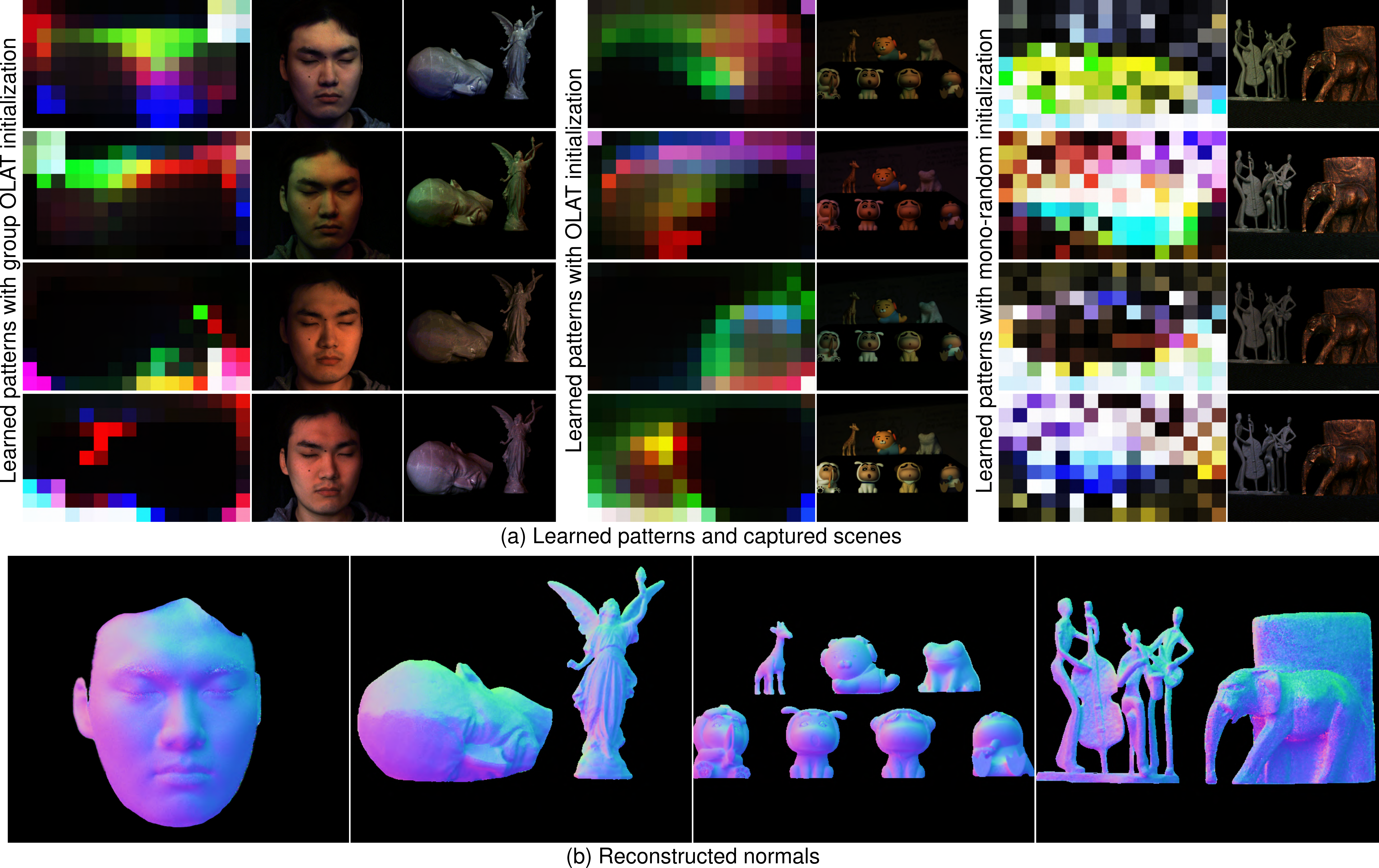}
		\caption{\textbf{Reconstruction results.} We reconstruct normals of diverse objects with the learned patterns using DDPS. Note that the patterns are learned on our 3D-printed training dataset.}
   		\label{fig:results}
 \vspace{-4mm}
 \end{figure*}

\paragraph{Real-world Objects}
For the experiments, we used 40 training {scenes} containing various 3D-printed objects. Even though increasing the number of training samples is feasible, we found that DDPS already allows for high-quality reconstruction for in-the-wild real-world objects in this configuration, as shown in Figure~\ref{fig:results}.
We speculate that this capability originates from effective rendering and analytical reconstruction without any additional training parameters as well as supervision with the 3D-printed dataset captured by a real setup. 

\begin{table}[!t]
\centering
\resizebox{0.9\columnwidth}{!}{
    \begin{tabular}{c|c|cc}
    \toprule[1pt]
    \multirow{1}{*}{Illumination }& \multirow{1}{*}{Number} & \multicolumn{2}{c}{Reconstruction error $\downarrow$} \\ \cline{3-4}
     patterns & of patterns & Initial  & Learned  \\ \hline \hline
        OLAT & 4 & {0.1707} & 0.0486 \\
        Group OLAT & 4 & 0.0805 & 0.0475 \\
        Mono-gradient & 4  & 0.0913 & {0.0443} \\ 
        Mono-complementary & 4 & 0.1044 & 0.0453 \\
        Tri-gradient & 2 & 0.0933 & 0.0512 \\
        Tri-complementary & 2 & 0.0923 & 0.0478\\ \hline
        Flat gray & 4 & 0.3930 & 0.0466 \\
        Mono-random & 4 & 0.2533 & 0.0484 \\
        Tri-random & 2 & 0.1461 & 0.0476\\ 
    \bottomrule[1pt]
    \end{tabular}
}
\caption{Comparison of display patterns without and with our end-to-end optimization. }
\label{tab:learned_illum}
\vspace{-1mm}
\end{table}

\paragraph{Number of Patterns}
Since photometric stereo solves for five unknowns (RGB diffuse albedo and surface normals), the minimum number of patterns is set to two, providing six measurements with the RGB channel for each. 
Table~\ref{tab:num_illum} shows that using two patterns learned by DDPS already outperforms any tested heuristic design using four patterns, demonstrating improved capture efficiency.
Moreover, using two learned patterns is often sufficient, as shown by the converged reconstruction errors. 

\begin{table}[!t]
\centering
\setlength{\tabcolsep}{25pt}
\resizebox{0.9\columnwidth}{!}{
    \begin{tabular}{c|cc}
    \toprule[1pt]
     \multirow{1}{*}{Number} & \multicolumn{2}{c}{Reconstruction error $\downarrow$} \\  \cline{2-3} of patterns & Initial  & Learned  \\ 
     \hline \hline
         2 & 0.1461 & 0.0476 \\
          3 & 0.1415 & 0.0467 \\
         4  & 0.1096 & {0.0463} \\ 
         5 & {0.1001} & 0.0467 \\ 
    \bottomrule[1pt]
    \end{tabular}
}
\caption{Quantitative results of reconstructed surface normals with varying number of patterns for the trichromatic random patterns.}
\label{tab:num_illum}
\vspace{-4mm}
\end{table}

\paragraph{Robustness to Simplifications}
For efficient end-to-end pattern learning, DDPS has made assumptions including light source modeling and intensity falloff in its image formation and reconstruction.
While the validity of these assumptions is often critical for conventional approaches that use synthetic training data, {DDPS exhibits robustness against such simplifications, as demonstrated in all the qualitative and quantitative results.} 
This is because the learned display patterns are optimized to achieve accurate normal reconstruction on {a real-world 3D-printed dataset, taking into account such assumptions.} 

\begin{figure}[t]
    \centering
    \includegraphics[width=\linewidth]{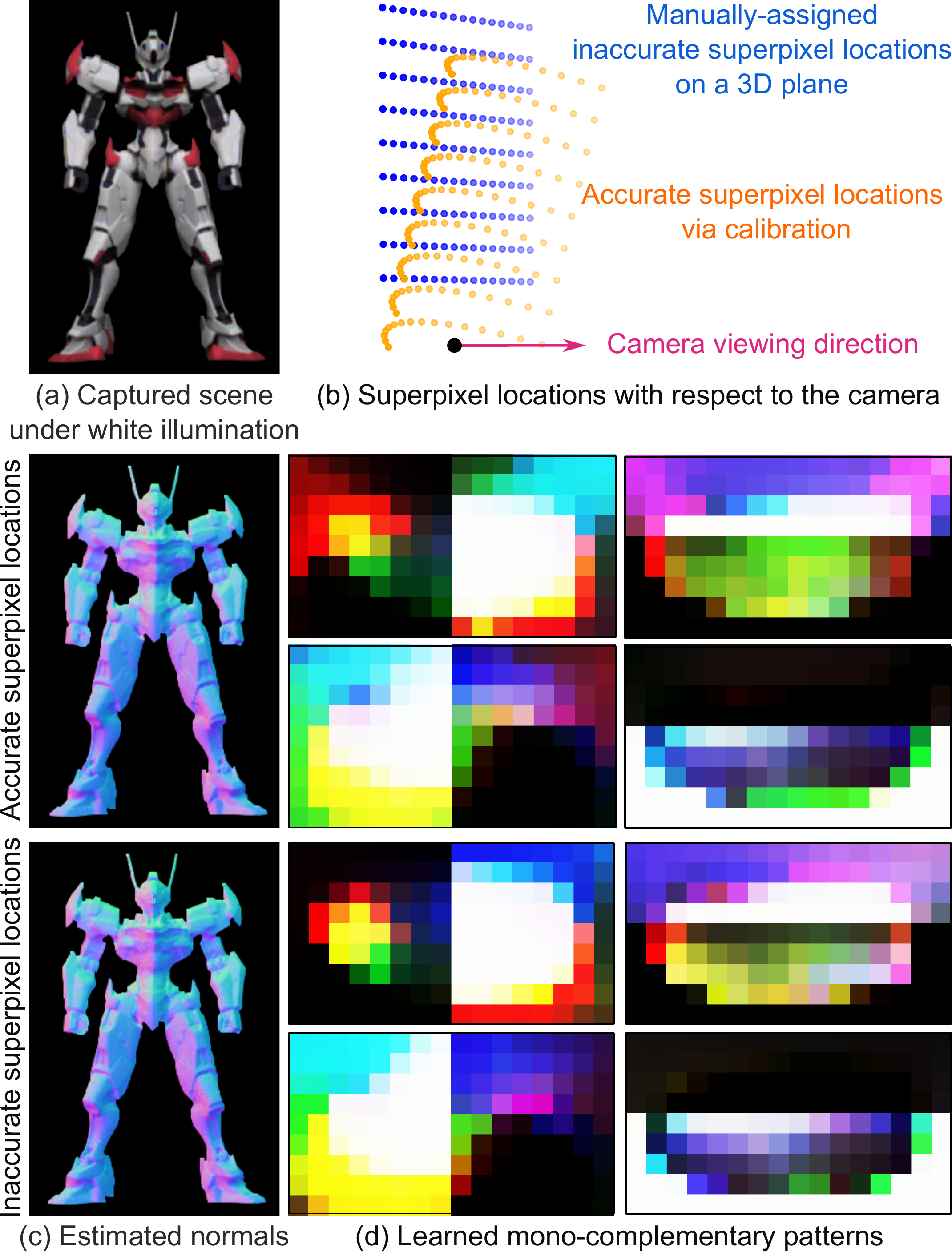}
    \caption{\textbf{Robustness against inaccurate superpixel locations.}
    We test DDPS for our calibrated curved-monitor superpixel locations, shown as (b) orange dots, and for the manually placed inaccurate plane superpixel locations, shown as blue dots, respectively.
    DDPS automatically compensates for the location error of the superpixels by (d) learning display patterns for such configuration, resulting in (c) high-quality normal maps. 
    }
    \label{fig:superpixel_cal}
\vspace{-3mm}
\end{figure}

Here, we conduct additional experiments to test the robustness of DDPS.
First, we evaluate DDPS under inaccurate superpixel locations. 
Instead of using our mirror-based calibration (Section~\ref{sec:system}), we manually place superpixels to lie at grid locations on a 3D plane, which deviates from the ground-truth locations. See Figure~\ref{fig:superpixel_cal}.
DDPS with the inaccurate superpixel locations still provides accurate normal reconstruction with the error {0.0456} comparable to {0.0453} corresponding to using accurate superpixel locations. 
Second, we evaluate the assumption of consistent intensity with respect to distance. DDPS {with and without} intensity fall-off show comparable reconstruction errors of {0.0429} and {0.0453}, indicating the robustness of DDPS against light fall-off modeling.
Third, we test test DDPS for an object at varying depths: 40/50/80/100\,cm.
Even though we assume planar scene geometry at a fixed distance of 50\,cm in our image formation, DDPS enables accurate normal reconstruction with the corresponding errors of {0.0494/0.0417/0.0428/0.0561} for the varying depths. 
That is, in that depth range, we achieve reconstruction errors lower than 0.0805, which is the error using the best-performing heuristic pattern, group OLAT for the 50cm-distant objects.
These experiments further demonstrate the robustness of DDPS against various simplifications. 

\vspace{-6mm}

\begin{figure}[!t]
    \centering
    \includegraphics[width=0.99\linewidth]{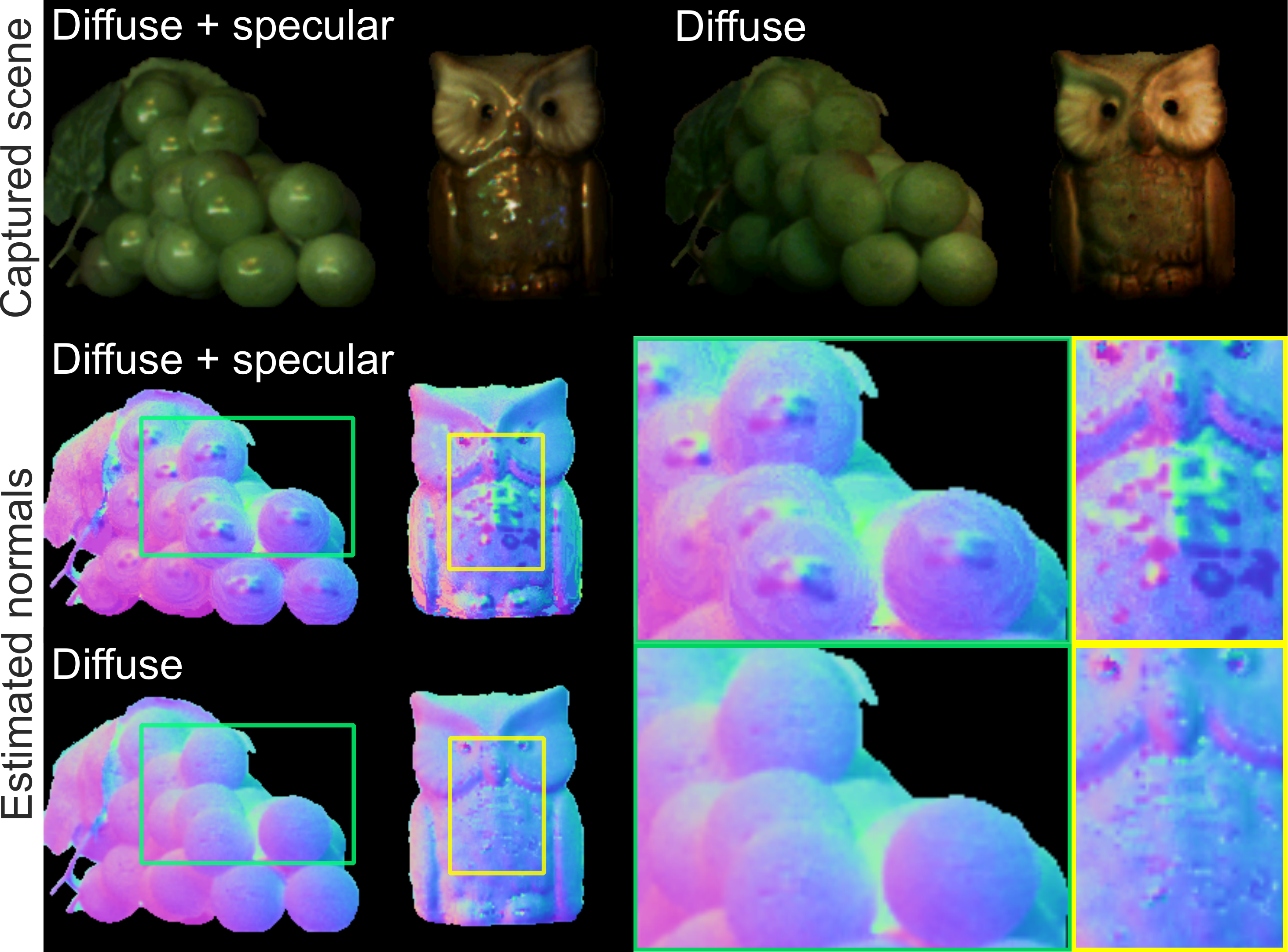}
    \caption{\textbf{Impact of diffuse-specular separation.} DDPS exploits polarization for optical diffuse-specular separation, leading to accurate normal reconstruction. 
    } 
    \label{fig:diffuse_specular}
\vspace{-4mm}
\end{figure}

\begin{figure}[]
    \centering
    \includegraphics[width=\linewidth]{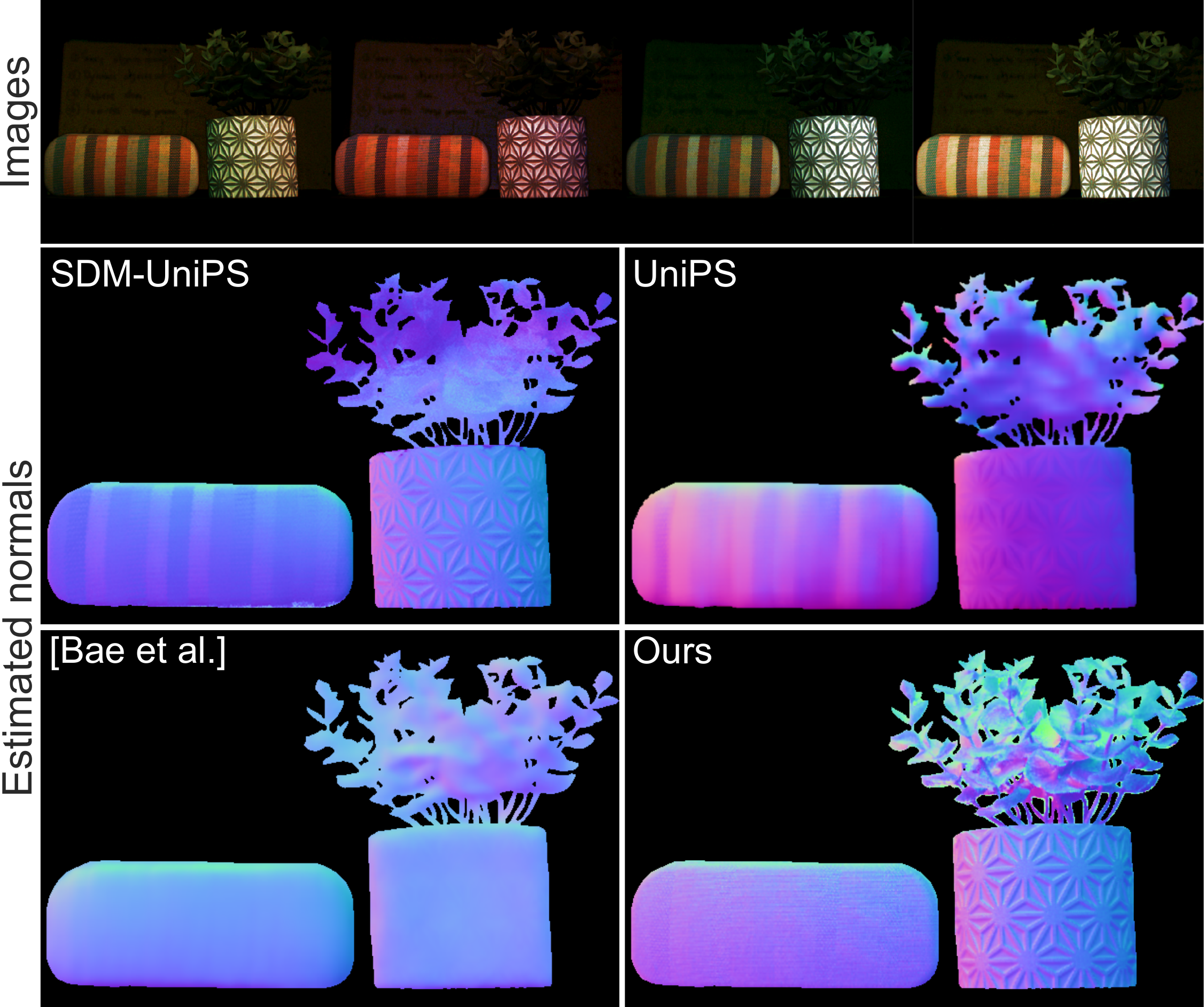}
    \caption{
    \textbf{Comparison to learning-based methods.} DDPS with the analytic reconstructor shows fine geometric details on the leafs, vase, and textile, outperforming the other methods.
    }
    \label{fig:comparison}
\vspace{-4mm}
\end{figure}

\paragraph{Impact of Diffuse-specular Separation}
In order to acquire diffuse-dominant images, DDPS exploits linearly-polarized light emitted from the monitor and the polarization camera.
Figure~\ref{fig:diffuse_specular} shows that the reconstructed surface normals from the diffuse-dominant images obtained by DDPS provide more accurate reconstruction than using the images containing both diffuse and specular reflections.
 
\paragraph{Comparison with Learning-based Photometric Stereo}
We compare the reconstructed normals using the learned patterns to state-of-the-art normal-reconstruction methods that leverage neural networks and support area light sources compatible with our learned patterns: UniPS~\cite{ikehata2022universal}, SDM-UniPS~\cite{ikehata2023scalable}, and Bae et al. \cite{bae2021estimating}. 
UniPS and SDM-UniPS use multiple images under diverse unknown illumination conditions. Bae et al. \cite{bae2021estimating} reconstruct the normal map from a single image.
Figure~\ref{fig:comparison} shows that DDPS outperforms the other methods. 
In particular, uncalibrated learned methods often fail to handle out-of-distribution examples such as the leaves in the scene.
In contrast, DDPS exploits shading cue for physically-valid and accurate normal reconstruction.

\paragraph{Learning-based Reconstructor}
DDPS uses analytic photometric stereo as a training-free and dataset-independent module for normal reconstruction. 
When we simply replace the analytic photometric stereo with a learning-based photometric stereo, UniPS~\cite{ikehata2022universal}, the average reconstruction error increases from 0.0475 to 0.0951, using the group-OLAT initialization. This degradation can be attributed to that the backward gradient for the display patterns does not flow as effectively due to the complex network structure of UniPS. 
Also, the network is not designed to effectively utilize complex display patterns. 
Developing a learning-based photometric stereo suitable for DDPS would be an interesting future work.
\section{Discussion}
\label{sec:discussion}
First, DDPS focuses on estimating normals, leaving depth reconstruction as a future work. Using multi-view cameras could resolve the problem and prompt research into optimizing patterns for multi-view cameras.
Second, we encountered challenges in achieving high-speed synchronization between the display and the camera. This could potentially be circumvented with external hardware triggering, which would facilitate the reconstruction of surface normals for dynamic objects.
Third, it would be interesting to apply DDPS for various types of display-camera systems such as a mobile phone.
Lastly, our image formation model does not consider shadow and  global illumination, which we further analyze in our Supplemental Document.
\section{Conclusion}
\label{sec:conclusion}

In this paper, we presented DDPS, a method for learning display patterns for robust display photometric stereo departing from using heuristic patterns.
Our differentiable framework consisting of basis-illumination image formation and analytic photometric stereo, the use of 3D printing for real-dataset creation, and display polarimetric separation allow for learning display patterns that leads to high-quality normal reconstruction for diverse objects.
Also, DDPS demonstrates robustness against various simplifications in image formation, reconstruction, and calibration. 
We believe that DDPS takes a step towards practical high-quality 3D reconstruction.
Beyond display photometric stereo, the principles underpinning DDPS would be applied to a range of illumination-camera systems, including light stages, mobile phones, and large-scale displays.

\paragraph{Acknowledgements}
This work was partly supported by Korea NRF (RS-2023-00211658, 2022R1A6A1A03052954, RS-2023-00280400), Samsung Advanced Institute of Technology, Samsung Research Funding \& Incubation Center for Future Technology grant (SRFC-IT1801-52), and Samsung Electronics.

\clearpage
{\small
\bibliographystyle{ieee_fullname}
\bibliography{main}
}

\end{document}


\begin{CJK}{UTF8}{}
\CJKfamily{mj}

\title{Differentiable Display Photometric Stereo \\
\it Supplemental Document 
}

\author{Seokjun Choi\footnotemark[2] ~ ~ ~
Seungwoo Yoon\footnotemark[2] ~ ~ ~
Giljoo Nam\footnotemark[1] ~ ~ ~
Seungyong Lee\footnotemark[2] ~ ~ ~
Seung-Hwan Baek\footnotemark[2] \\
\footnotemark[2]~ POSTECH   ~ ~ ~   ~ ~ ~  \footnotemark[1]~ Meta Reality Labs\\
}

\maketitle

\hypersetup{linkcolor=black}
\noindent
In this Supplemental Document, we provide additional results and details of DDPS.
\tableofcontents

\newpage

\section{Detailed Formulation of Photometric Stereo}
We provide the detailed formulations of the normal and albedo reconstruction as follows:
\begin{align}
\label{eq:photometric_stereo_rgb_matrix_normal}
\underbrace{
\scalebox{0.7}{
$\begin{bmatrix}
{I}^{R}_1 \\
\vdots\\
{I}^{R}_K \\
{I}^{G}_1 \\
\vdots\\
{I}^{G}_K \\
{I}^{B}_1 \\
\vdots\\
{I}^{B}_K
\end{bmatrix}$ }}_{\mathbf{I}} &\mathsmaller{=} 
\underbrace{
\scalebox{0.7}{
$\begin{bmatrix}
\rho^R \\
\vdots\\
\rho^R \\
\rho^G \\
\vdots\\
\rho^G \\
\rho^B \\
\vdots\\
\rho^B
\end{bmatrix}$ }
}_{\boldsymbol{\rho}}
\mathsmaller{\odot}
\underbrace{
\scalebox{0.7}{
$\begin{bmatrix}
\mathcal{M}^{R}_{1,1} & \cdots & \mathcal{M}^{R}_{1,P} \\
\vdots & \ddots & \vdots \\
\mathcal{M}^{R}_{K,1} & \cdots & \mathcal{M}^{R}_{K,P} \\
\mathcal{M}^{G}_{1,1} & \cdots & \mathcal{M}^{G}_{1,P} \\
\vdots & \ddots & \vdots \\
\mathcal{M}^{G}_{K,1} & \cdots & \mathcal{M}^{G}_{K,P} \\
\mathcal{M}^{B}_{1,1} & \cdots & \mathcal{M}^{B}_{1,P} \\
\vdots & \ddots & \vdots \\
\mathcal{M}^{B}_{K,1} & \cdots & \mathcal{M}^{B}_{K,P} 
\end{bmatrix}$ }
}_{\mathbf{M}}
\underbrace{
\left[ \begin{matrix}
l_{1,x} & l_{1,y} & l_{1,z}  \\
\vdots & \vdots & \vdots \\
l_{P,x} & l_{P,y} & l_{P,z}  \\
\end{matrix}\right]
}_{\mathbf{l}}
\underbrace{
\scalebox{0.7}{
$\begin{bmatrix}
N_x \\
N_y \\
N_z
\end{bmatrix}$ }
}_{\mathbf{N}}.
\end{align}
\begin{align}
\label{eq:photometric_stereo_rgb_matrix_diffuse_albedo}
\underbrace{
\scalebox{0.8}{
$\begin{bmatrix}
{I}^{c}_1 \\
\vdots\\
{I}^{c}_K \\
\end{bmatrix}$ }
}_{\mathbf{I}^c} &= 
\rho^c
\underbrace{
\scalebox{0.8}{
$\begin{bmatrix}
\mathcal{M}^{c}_{1,1} & \cdots & \mathcal{M}^{c}_{1,P} \\
\vdots & \ddots & \vdots \\
\mathcal{M}^{c}_{K,1} & \cdots &\mathcal{M}^{c}_{K,P} \\
\end{bmatrix}$ }
}_{\mathbf{M}^c}
\mathbf{l}\mathbf{N}.
\end{align}

\section{Details on Initial Patterns}
We utilize a variety of initialization patterns, each with its own characteristics:

\begin{itemize}
\item \textbf{OLAT}~\cite{sun2020light}: Each OLAT pattern consists of a boundary superpixel with an intensity value of 0.9, while the other superpixels have an intensity of 0.1.
\item \textbf{Group OLAT}~\cite{bi2021deep}: In this pattern, we use a group of $3\times3$ superpixels. Each pattern activates a different group superpixel.
\item \textbf{Monochromatic gradient}~\cite{ma2007rapid}: This pattern includes x- and y-gradient patterns in both forward and backward directions. The intensity values range from 0.1 to 0.9.
\item \textbf{Monochromatic complementary}~\cite{kampouris2018diffuse}: Similar to the monochromatic gradient pattern, this pattern includes x- and y-binary patterns in both forward and backward directions, with intensity values ranging from 0.1 to 0.9.
\item \textbf{Trichromatic complementary}~\cite{lattas2022practical}: This pattern involves using complementary x-binary patterns for the red channel, complementary y-binary patterns for the blue channel, and turning on different quadrants for the green channel.
\item \textbf{Trichromatic gradient}~\cite{meka2019deep}: This pattern is a modification of the trichromatic complementary pattern. It includes x-gradient patterns for the red channel, y-gradient patterns for the blue channel, and a gradient from the center to the boundary for the green channel.
\item \textbf{Monochromatic random}: Each superpixel intensity is randomly drawn from a uniform distribution between zero and one.
\item \textbf{Trichromatic random}: Similar to the monochromatic random pattern, each superpixel intensity for each color channel is randomly drawn from a uniform distribution between zero and one.
\item \textbf{Flat gray}: Each superpixel intensity is sampled from a Gaussian distribution with a mean of 0.5 and a standard deviation of 0.01.
\end{itemize}

Figure~\ref{fig:learned_illum}(a) shows the initial patterns. We set the minimum and maximum intensity values of initial patterns non-saturated from 0.1 to 0.9, to avoid zero gradient in end-to-end optimization.
\section{Additional Analysis on Learned Patterns}

\begin{figure*}[t]
    \centering
    \includegraphics[width=\linewidth]{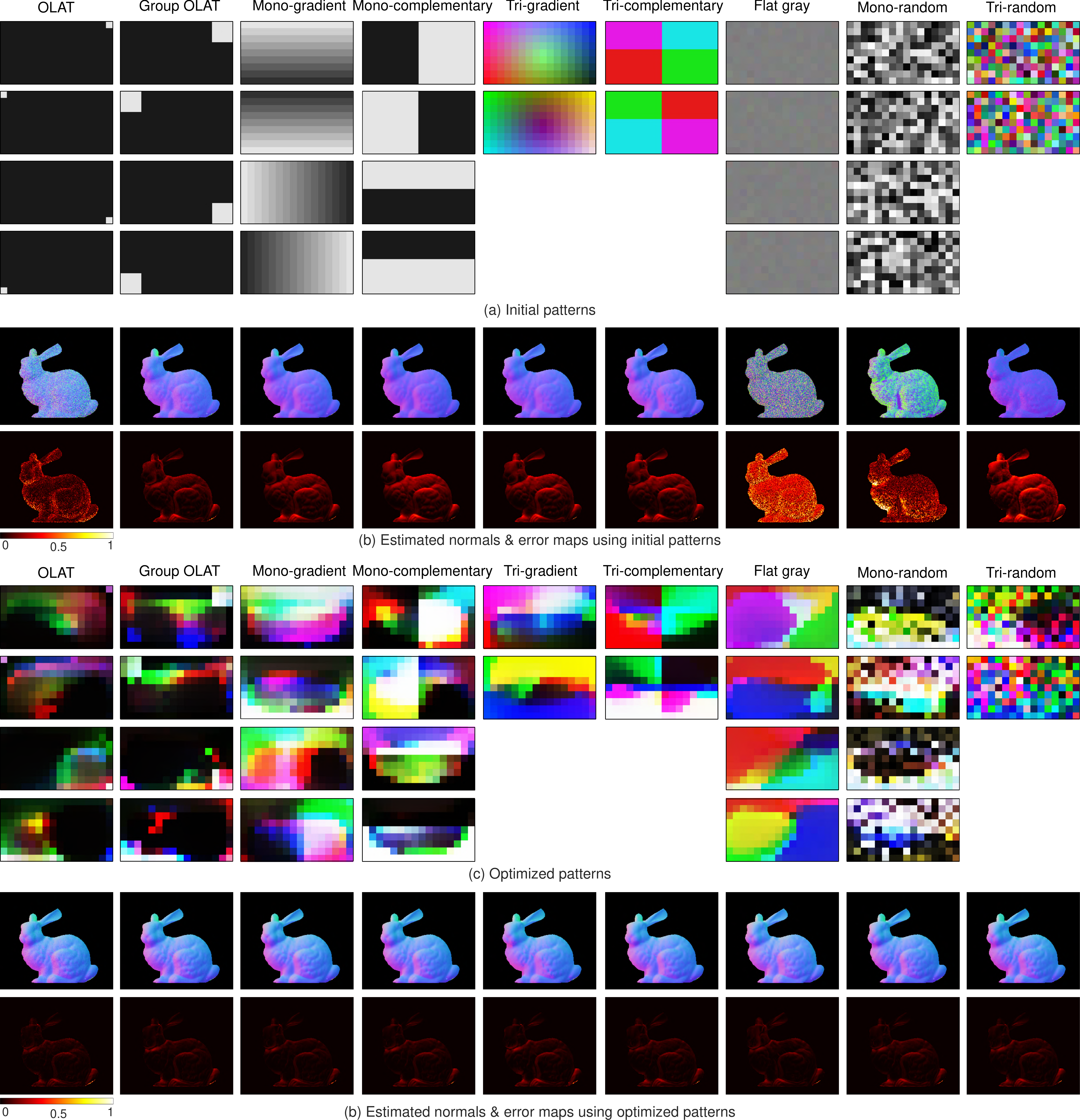}
    \caption{\textbf{Learned patterns.} (a) Heuristically-designed display patterns results in (b) sub-optimal normal reconstruction. (c) DDPS allows for learning display patterns, leading to (d) high-quality normal reconstruction.  
    } 
    \label{fig:learned_illum}
\end{figure*}

Figure~\ref{fig:learned_illum}(c) illustrates the illumination patterns learned using DDPS with every initialization pattern. 
DDPS consistently improves reconstruction quality compared to initial patterns, indicating that heuristically-designed patterns can be further optimized for specific display-camera configurations.
We note that the overall shape of the patterns tends to be determined during the early stages of the training process.
We refer to the Supplemental Video for the progression of pattern learning.

\clearpage

\clearpage

\section{Additional Analysis on the Number of Illumination Patterns}
We show the impact of using varying numbers of illumination patterns for flat-gray and trichromatic-random patterns ranging from two to five. The reconstruction results on the test dataset of 3D-printed objects are presented in Table~\ref{tab:num_illum} computed with $\texttt{loss}(\cdot)$.

\begin{table}[h]
\centering
\resizebox{0.5\columnwidth}{!}{
    \begin{tabular}{c|c|cc}
    \toprule[1pt]
    \multirow{1}{*}{Illumination }& \multirow{1}{*}{Number} & \multicolumn{2}{c}{Reconstruction error $\downarrow$} \\ \cline{3-4}
     patterns & of patterns & Initial  & Learned  \\ \hline \hline
        Tri-random & 2 & 0.1461 & 0.0476 \\
         Tri-random & 3 & 0.1415 & 0.0467 \\
         Tri-random& 4  & 0.1096 & {0.0463} \\ 
        Tri-random & 5 & {0.1001} & 0.0467 \\ \hline
        Flat gray & 2 & 0.3549 & 0.0614 \\
         Flat gray & 3 & 0.4007 & 0.0469 \\
         Flat gray& 4  & 0.3930 & 0.0466 \\ 
        Flat gray & 5 & 0.4049 & 0.0462 \\ 
        \bottomrule[1pt]
    \end{tabular}
}
\caption{Quantitative results of reconstructed surface normals with varying number of patterns for the trichromatic random patterns. }
\label{tab:num_illum}
\end{table}

\section{Details on Capture System}
For the display, we use a commercial large curved LCD monitor (Samsung Odyssey Ark).
The monitor has a {55''} liquid-crystal display with 2160$\times$3840 pixels, peak brightness of 1000 $cd/m^2$. 
Each pixel of the monitor emits horizontally linearly-polarized light at trichromatic RGB spectra due to the polarization-sensitive optical elements of LCD. 
We use a polarization camera (FLIR BFS-U3-51S5PC-C) with on-sensor linear polarization filters at four different angles. Thus the polarization camera captures four linearly-polarized light intensities at the angles $0^\circ, 45^\circ, 90^\circ, 135^\circ$ as $I_{0^\circ}$, $I_{45^\circ}$, $I_{90^\circ}$, $I_{135^\circ}$.
{Instead of using an expensive polarization camera, adopting a conventional camera with linear-polarization film is one affordable alternative. 
Perpendicular polarization axis of the film to the display enables capturing diffuse images.}

\paragraph{Device Control}
To control the display patterns and operate the polarization camera, we use the PyGame and PySpin libraries, respectively. The devices are connected to a desktop computer via an HDMI cable and a USB3 cable. Our setup employs software synchronization between the display and the camera. 

\section{Iterative Normal-albedo Reconstruction}

Once the surface normal $\mathbf{N}$ is obtained, we rewrite the previous Equation~\eqref{eq:photometric_stereo_rgb_matrix_diffuse_albedo} in the main paper to solve for the albedo again:
\begin{align}
\label{eq:photometric_stereo_rgb_matrix_diffuse_albedo_rewrite}
\mathbf{I}^c=\rho^c \odot \mathbf{M}^c \mathbf{l} \mathbf{N},
\end{align}
where {$\mathbf{I}^c \in \mathbb{R}^{K \times 1}$, $\mathbf{M}^c \in \mathbb{R}^{K\times P}$} are the per-channel versions of the original vector $\mathbf{I}$ and matrix $\mathbf{M}$.
For each channel $c\in \{R,G,B\}$, we estimate the albedo {$\rho^c \in \mathbb{R}$} using the pseudo-inverse method as ${\rho^c} \leftarrow \mathbf{I}^c\left( \mathbf{M}^c \mathbf{l} \mathbf{N} \right)^{\dag}$.
We could iterate the normal estimation and the albedo estimation further for higher accuracy, which we found produces marginal improvements in the reconstruction quality. 

We evaluate our normal-albedo reconstruction methodology iteratively on initial patterns, using the estimated albedo to calculate the subsequent normal. Our tests reveal normal-reconstruction errors of 0.0805, 0.0798, and 0.0798 for the zero-iteration, first-iteration, and second iteration respectively. These results display negligible difference, signifying that additional iterations do not significantly improve the accuracy of normal reconstruction. Consequently, for the sake of computational efficiency, we have chosen to implement a single-stage reconstruction process.

\section{Optimization Details}
We use a batch size of 2 and a learning rate of {0.3} with a learning-rate decay rate of {0.3} and a step size of {5 epoch}.
We run the training process for {30} epochs, which takes 15 minutes on a single {NVIDIA GeForce RTX 4090} GPU.

\section{Calibration Details}
\subsection{Mirror-based Geometric Calibration} 
We propose a mirror-based calibration method for estimating the intrinsic parameter of the camera and the location of each pixel of the monitor with respect to the camera. Figure~\ref{fig:calib}(a) illustrates our geometric calibration.

We first print a checkerboard on a paper. Then, we place a planar mirror at a certain pose in front of the camera while displaying a grid of white pixels on the monitor. We then capture the mirror that reflects some of the grid points, to which the corresponding monitor pixel coordinates are manually assigned. Next, we put the printed checkerboard on top of the planar mirror and capture another image, which now contains the checkerboard. We repeat this procedure by varying poses of the planar mirror, resulting in multiple pairs of a checkerboard image and a mirror image reflecting grid points.

From the checkerboard images, we estimate the intrinsic parameter of the camera and the 3D pose of each checkerboard~\cite{zhang2000flexible}. 
We then detect the 3D points of the grid points in each mirror image with the known size of the monitor and obtain the 3D points of intermediate monitor pixels via interpolation.

\subsection{Radiometric Calibration}
\begin{figure*}[h]
    \centering
    \includegraphics[width=\textwidth]{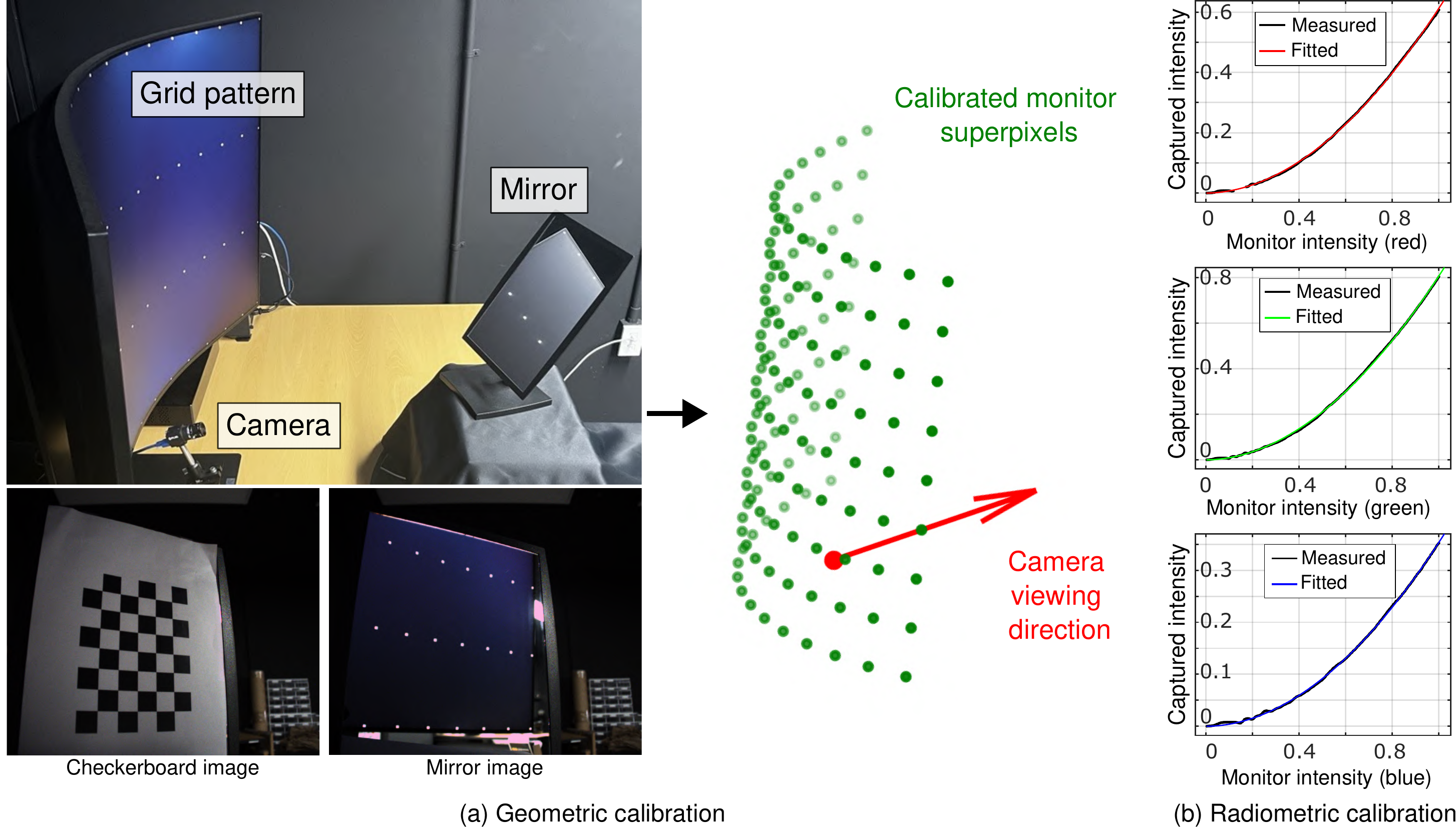}
    \caption{\textbf{Calibration.} (a) We calibrate the parameters of the camera and monitor using a mirror that reflects grid display patterns. (b) We also calibrate the non-linear mapping of monitor pixel values to emitted radiance for each color channel. 
    }
    \label{fig:calib}
\end{figure*}

The emitted radiance from the monitor does not have a linear relationship with the pixel values of the display pattern. To account for this nonlinearity, we capture images of gray patches on a color checker under different intensity values of the display pattern. 
We then fit an exponential function to the captured intensity values with respect to the monitor pixel values for each color channel.
Figure~\ref{fig:calib}(b) shows the fitted curves.

\clearpage

\section{Example of Stokes-vector Reconstruction}
\begin{figure*}[h]
\centering
\includegraphics[width=\linewidth]{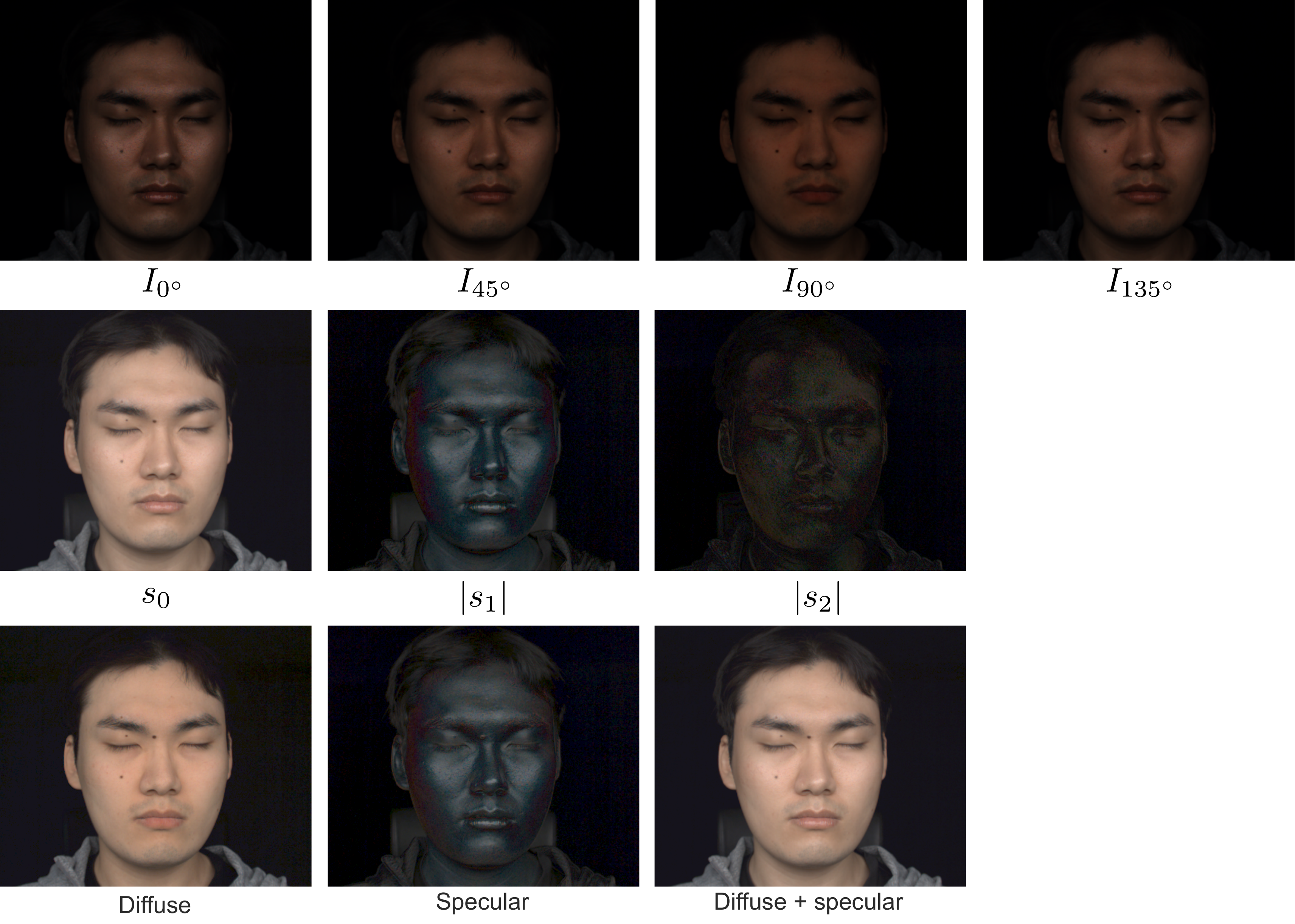}
\caption{\textbf{Stokes-vector and diffuse-specular separation.} The linearly-polarized images  $I_{0^\circ}$, $I_{45^\circ}$, $I_{90^\circ}$, $I_{135^\circ}$, Stokes-vector elements $s_0$, $s_1$, $s_2$, diffuse reflection $I$, specular reflection $S$, and diffuse-specular reflection}
\label{fig:stokes}
\end{figure*}

To separate the diffuse and specular images, we reconstruct Stokes vector  as intermediate results. Figure~\ref{fig:stokes} shows the linearly-polarized images $I_{0^\circ}$, $I_{45^\circ}$, $I_{90^\circ}$, $I_{135^\circ}$, Stokes-vector elements $s_0$, $s_1$, $s_2$, diffuse reflection $I$, specular reflection $S$, and diffuse-specular reflection.

\clearpage
\section{Dataset}

\subsection{Training and Testing Sets}
Figures~\ref{fig:training_dataset} and \ref{fig:test_dataset} shows the datasets used for training and testing, respectively.

\subsection{3D Printing and 3D Models}
In our work, we use an FDM-based 3D printer due to its affordability and the diversity it offers. This type of 3D printer can utilize filaments of various textures, colors, and materials, thereby enhancing the diversity of our models. The 3D models are converted into gcode through a slicing process for 3D printing. It is worth noting that other types of 3D printers, such as SLA, SLS, and DLP, could further diversify the dataset.
We 3D-print {11} different 3D models using a FDM-based 3D printer (Anycubic Kobra) that has a printing resolution of $\sim$0.2\,mm. We use multiple filaments (PLA, PLA+, Matte PLA, eSilk-PLA, eMarble-PLA, Gradient Matte PLA, PETG) that provide diverse appearances in terms of color, scattering, and diffuse/specular ratios. The 3D-printed objects have volumes ranging from $198.9\,\mathrm{cm}^3$ to $3216.423\,\mathrm{cm}^3$. 
Our dataset includes 3D models, comprising busts, animal figures, and character models. These models were chosen for their diverse geometric features and asymmetry, which aids pose estimation. We foresee the potential for expanding the diversity of our models by leveraging large public 3D model datasets.

\subsection{Pose Estimation and Normal Rendering}
For constructing a dataset of 3D-printed objects, images are taken by the calibrated camera, under the basis illumination which is a white square on a portion of the monitor screen. With the fixed object, we took photographs of the object under a total of 144 basis illuminations, and by compositing photographs through relighting, one can synthesize a photograph of an object taken under arbitrary light sources. For ease of later pose estimation, the backgrounds of the captured images must be removed for which we adopt Adobe Photoshop for the background removal.

\begin{figure*}[t]
\centering
\includegraphics[width=\textwidth]{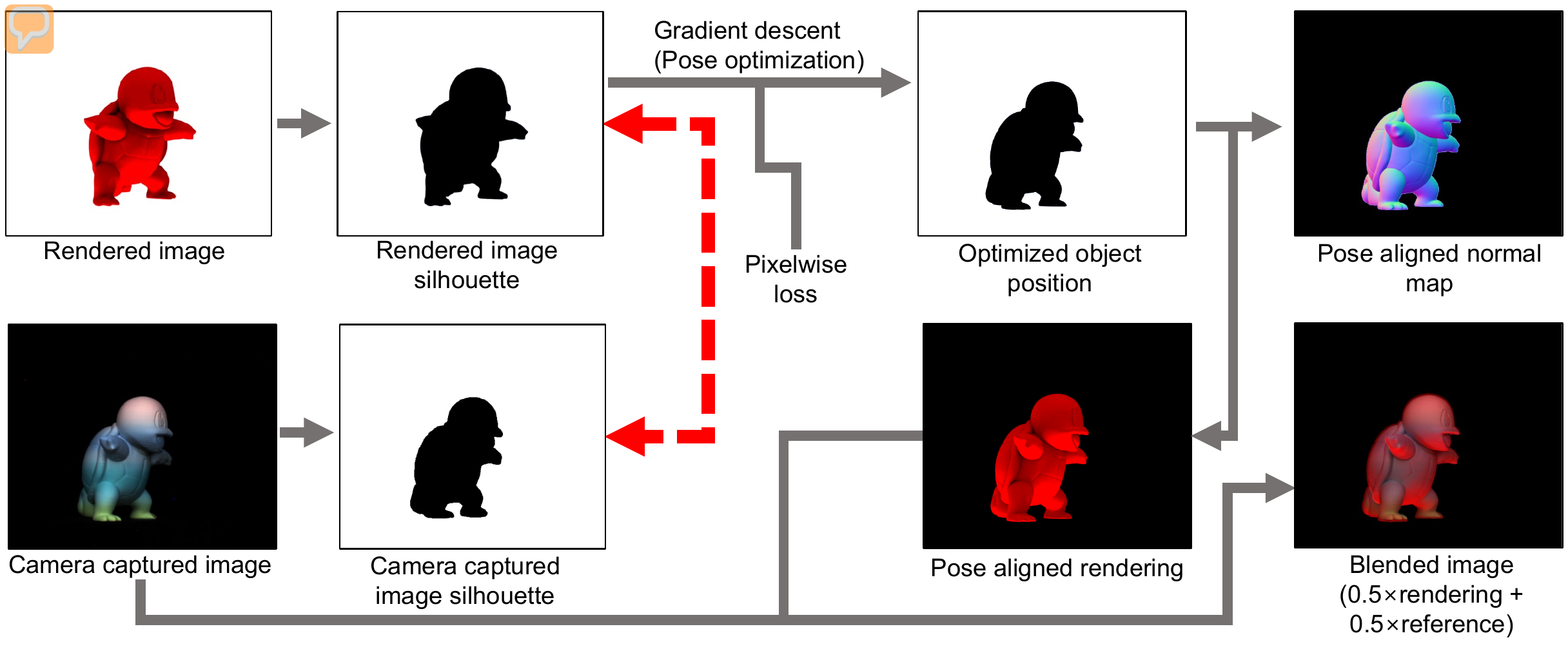}
\caption{\textbf{The process of pose alignment.} We optimize the object position parameters using pixel-wise L2 loss between the rendered image and real-world image silhouettes. As shown in the blended image, the pose estimation process well aligns the object with the reference image, ensuring a proper correspondence between the two.}
\label{fig:pose_estimation_process}
\end{figure*}

Even though we possess both real-world images and precise 3D model information of the objects, we need to align the object in the image with the corresponding 3D model by minimizing the reprojection error. To this end, we render silhouettes of objects using 3D mesh information and object position parameters. Then calculate the pixel-wise MSE loss between the silhouette image of the photograph and the rendered one. We used silhouettes instead of RGB rendering because it is challenging to exactly reproduce the RGB intensity given unknown reflectance.
The acquired position parameters and 3D model information are used as scene parameters, and we use the normal rendering functions provided by Mitsuba3.
The overall process of dataset generation is shown in Figure~\ref{fig:pose_estimation_process}.

\begin{figure*}[t]
\centering
\includegraphics[width=\textwidth]{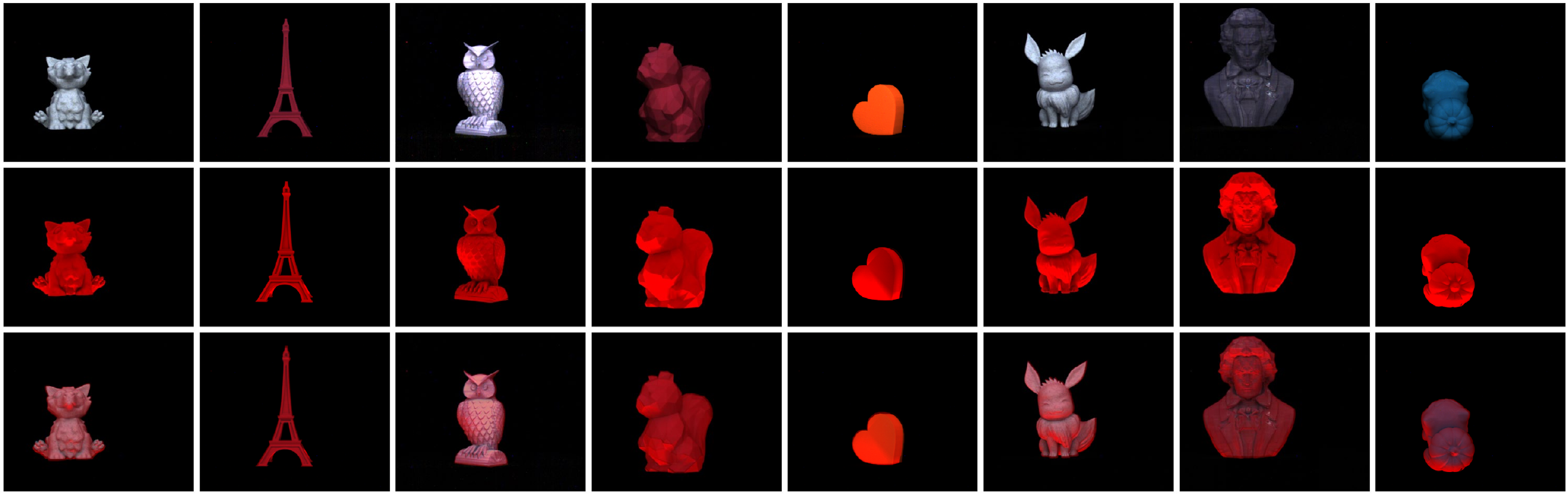}
\caption{\textbf{A qualitative visual representation of the pose estimation results.} The first row shows captured images, the second row represents rendered images with optimized poses, and the images in the third row are blended ones which are the equally weighted sum of images in the first and second rows.}
\label{fig:pose_estimation_results}
\end{figure*}

\begin{figure*}[t]
\centering
\includegraphics[width=\textwidth]{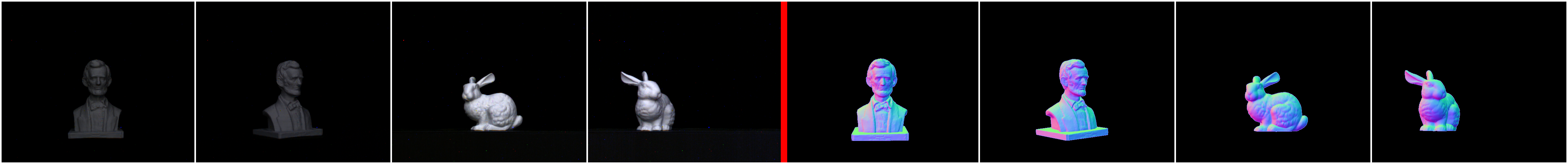}
\caption{\textbf{Visualization of our test dataset.} Captured images are on the left and their corresponding normal maps are on the right.}
\label{fig:test_dataset}
\end{figure*}

\begin{figure*}[t]
\centering
\includegraphics[width=\textwidth]{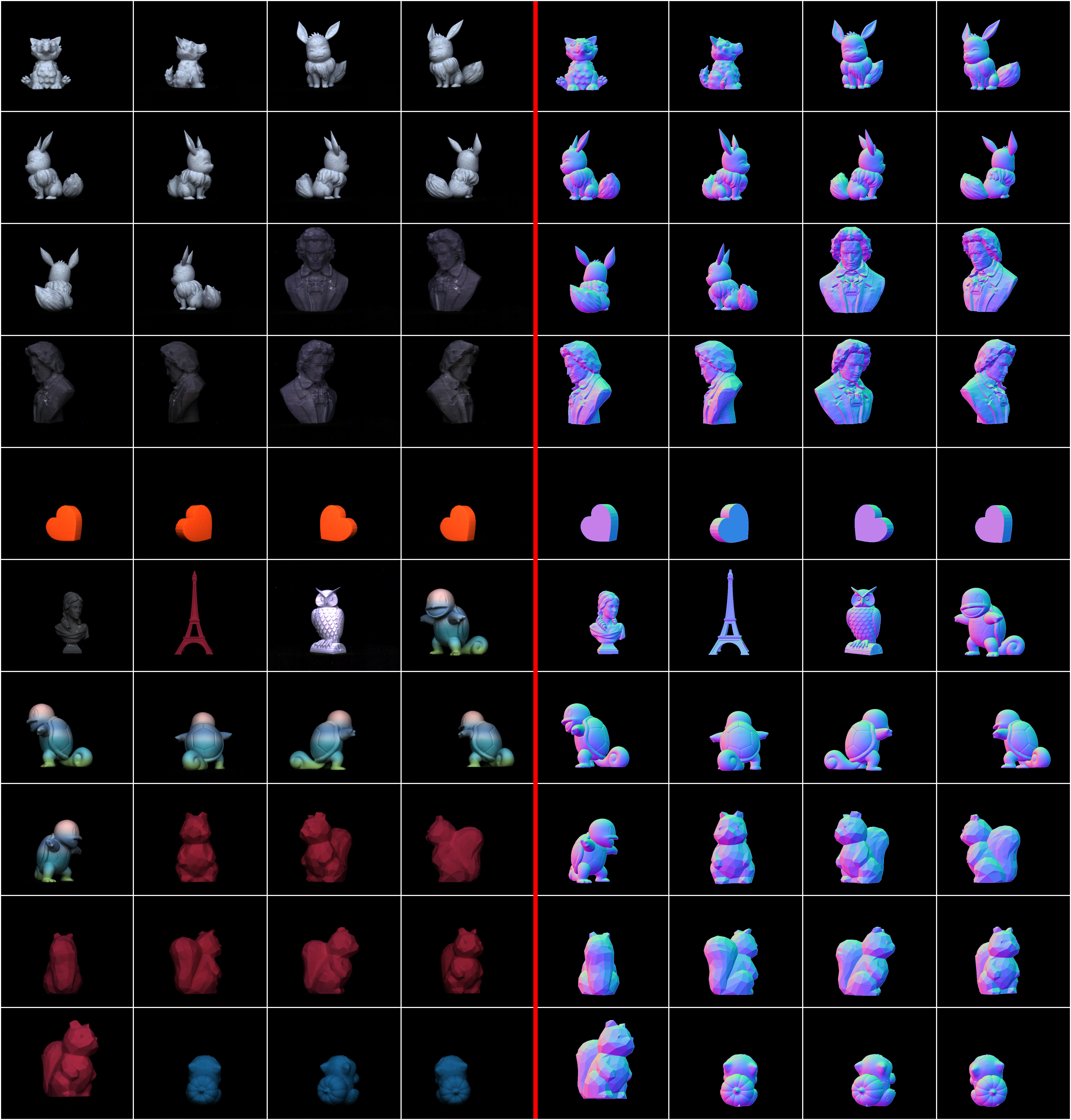}
\caption{\textbf{Visualization of our training dataset.} Images on the left are captured images and on the right are their ground-truth normal maps.}
\label{fig:training_dataset}
\end{figure*}

The pixel-wise mean squared error value is within the range of 0.0015 to 0.0028, depending on the size of the object in the image and the background removal. This low loss value signifies that the pose estimation is accurate, thereby confirming that the dataset offers a sufficiently precise representation to be deemed as ground-truth data. Figure~\ref{fig:pose_estimation_results} presents the qualitative results of the pose estimation accuracy.  Figure~\ref{fig:training_dataset} and Figure~\ref{fig:test_dataset} show our training and test dataset respectively.

\subsection{Sub-milimeter 3D-printing Error}
\begin{figure*}[h]
\centering
\includegraphics[width=0.7\textwidth]{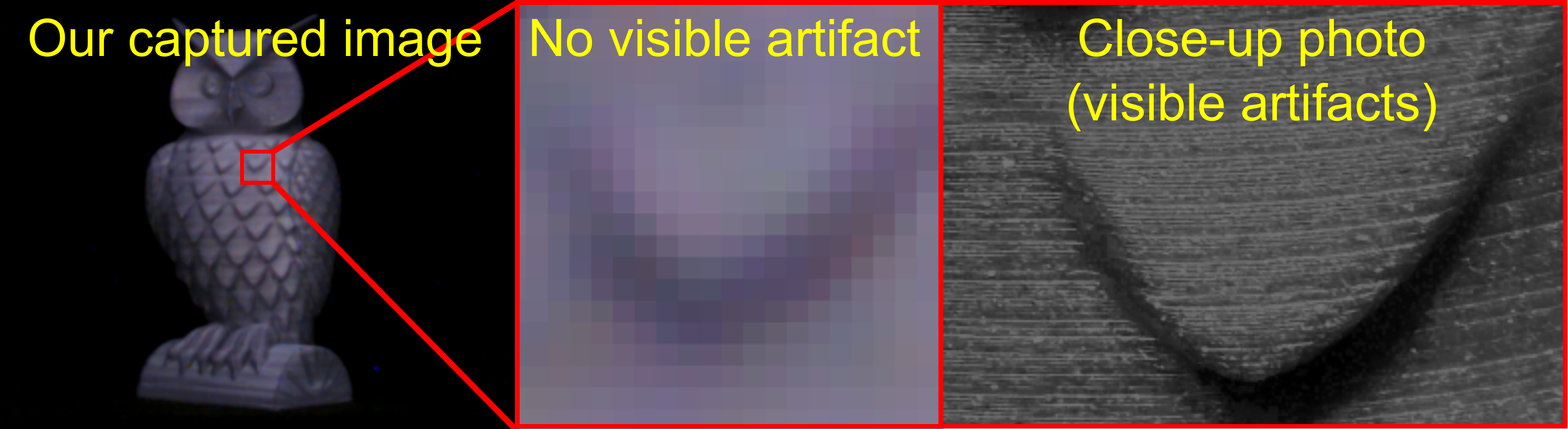}
\caption{\textbf{3D-printing error.} 3D-printing artifact is too small to be observed in our captured image. It can be observed in a close-up photo.}
\label{fig:submilimeter}
\end{figure*}

{Our 3D printer has 0.2mm resolution. Figure~\ref{fig:submilimeter} shows that captured images do not present visible artifacts, which is attributed to target distance, camera FoV, and lens blur.
Also, assuming a zero-mean distribution for the error, it would be canceled out as high-frequency noise during optimization.}

\clearpage

\clearpage

\section{Additional Discussion} 
\subsection{Dynamic Objects}
\label{sec:dynmaic_objects}
Our current experimental prototype supports software synchronization, which limits the operation speed of display-camera capture.
Here, hardware synchronization would unlock the full potential of the display-camera setup by supporting the maximum framerate of the devices, reaching over 150\,fps.
This may require a hardware trigger mechanism between the camera, display, and GPU, which we exclude from our scope that focuses on the methodology of DDPS.

If the hardware synchronization can be implemented, we could capture polarimetric images under repeating $N$ different monitor patterns at a framerate of 150 FPS for both display and imaging. Per each frame, we perform diffuse-specular separation and obtain diffuse image $I_i$ for the $i$-th monitor pattern. This results in a duration of $1/(15N)$ seconds for capturing a scene under $N$ patterns, which assumes marginal object movements during the capture time. Using optical flow may resolve minor alignment problem.
Specifically, at any input frame, we gather $N-1$ neighboring frames, from which surface normals and diffuse albedo could be estimated by our photometric stereo method.

\subsection{Superpixels}
\begin{figure}[h]
\centering
\includegraphics[width=0.7\linewidth]{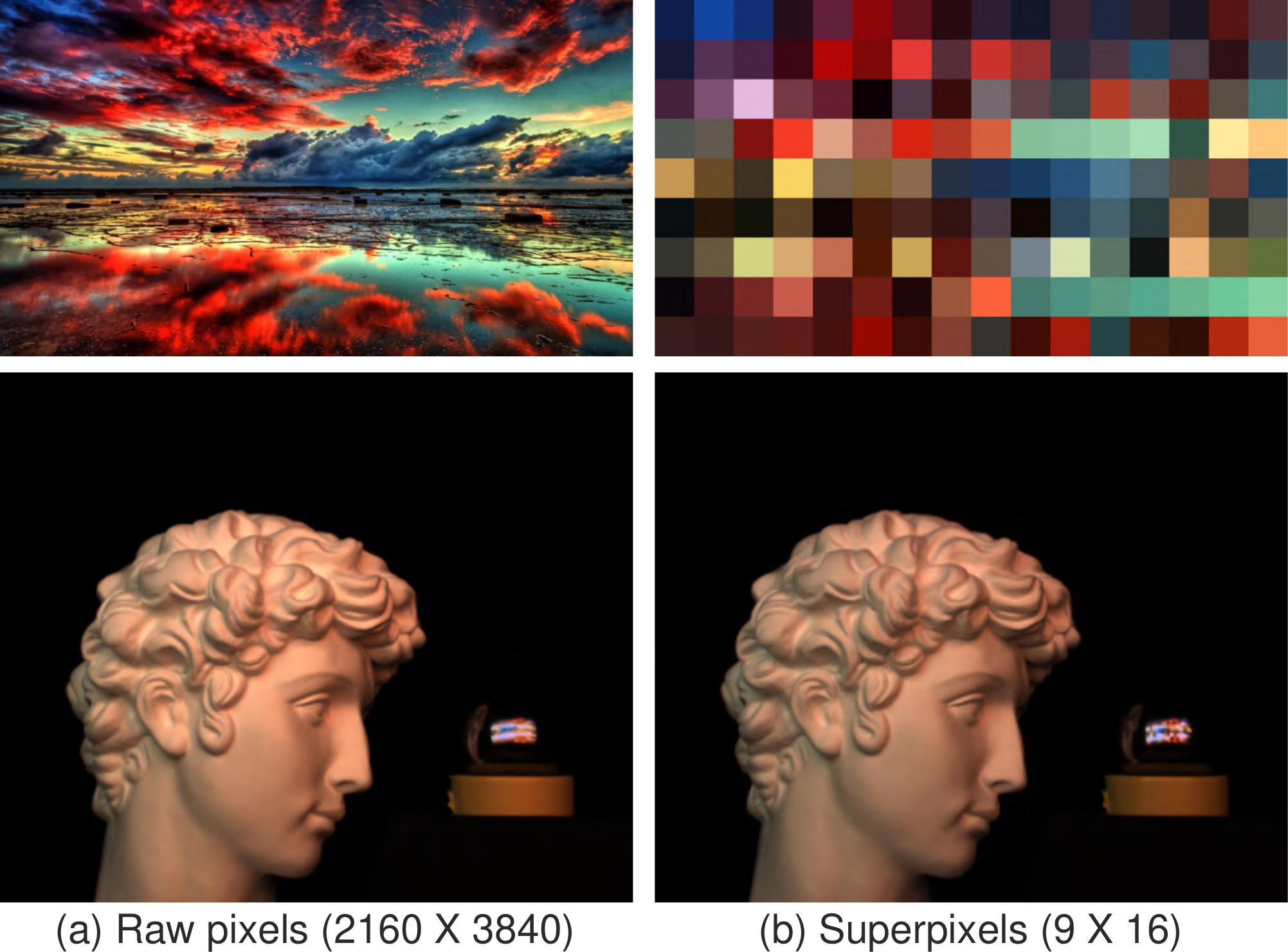}
\caption{\textbf{Comparison on different resolution of illumination.} We compare two images under (a) the illumination of raw display resolution (2160$\times$3840), and (b) the downsampled illumination with superpixels (9$\times$16). }
\label{fig:superpixel}
\end{figure}

We opt to use superpixels instead of raw pixels from the display for computational efficiency.
Figure~\ref{fig:superpixel} illustrates the similarity between the captured images when displaying a natural image using the raw display resolution and the downsampled version with superpixels.
{The low-frequency characteristics of projected illumination allow for using a low superpixel resolution.}
Although using more pixels for DDPS may enhance reconstruction accuracy by learning fine-grained patterns, the use of superpixels strikes a balance between computational efficiency and sufficient representation of the display. 
{This is essential since GPU memory must accommodate the image formation, reconstruction, and optimization of the display patterns. Using 9$\times$16 superpixels costs 12 GB memory.  
We confirmed that using 4$\times$8 superpixels results in reconstruction error of 0.0658 comparable to 0.0443 of the 9x16 setup. 
We used the learned patterns initialized with four mono-gradient patterns.
{Exploiting native 8M display pixels leaves as an interesting future work for inverse rendering where high-frequency cue is needed.}
}

\clearpage

\subsection{Global Illumination}
\label{global_illumination}
\begin{figure*}[h]
\centering
\includegraphics[width=0.9\textwidth]{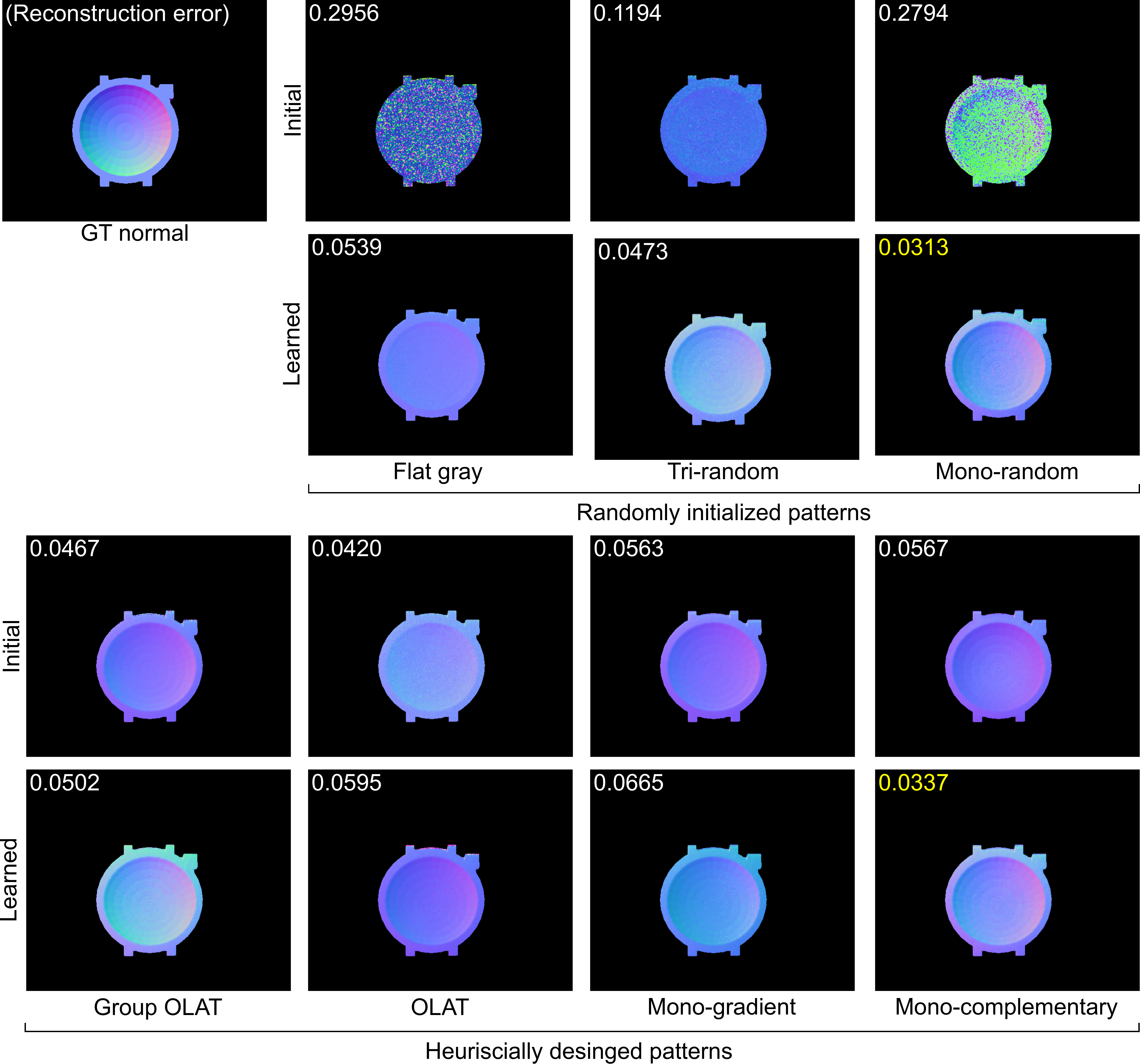}
\caption{\textbf{Normal reconstruction on a concave bowl.} Reconstruction often fails with a concave bowl. However, some patterns like mono-random and mono-complementary shows robust reconstruction results.}
\label{fig:global_illumination}
\end{figure*}

Our image formation model and reconstruction method do not consider global illumination, and also our training dataset does not contain objects that incur strong global illumination. 
As such, DDPS fails handling objects with significant interreflections.
We compare DDPS using initial patterns versus optimized patterns on a concave bowl.
Figure~\ref{fig:global_illumination} shows the reconstruction results and quantitative reconstruction losses with various initial/optimized patterns.
The heuristic patterns (e.g., group OLAT, OLAT, monochromatic gradient) estimates more accurate normals than optimized ones.
This is because the sparse initial-heuristic patterns result in less-pronounced inter-reflections.
OLAT patterns show trade-off between accurate reconstruction and noisy result due to the sparsity.
Some cases such as monochromatic complementary and monochromatic random patterns shows robust reconstruction on a concave bowl.
This demonstrates that DDPS could potentially be improved to handle the global illumination case.

\clearpage
\subsection{Optimal Lighting versus Random Lighting}
\begin{figure*}[h]
\centering
\includegraphics[width=\linewidth]{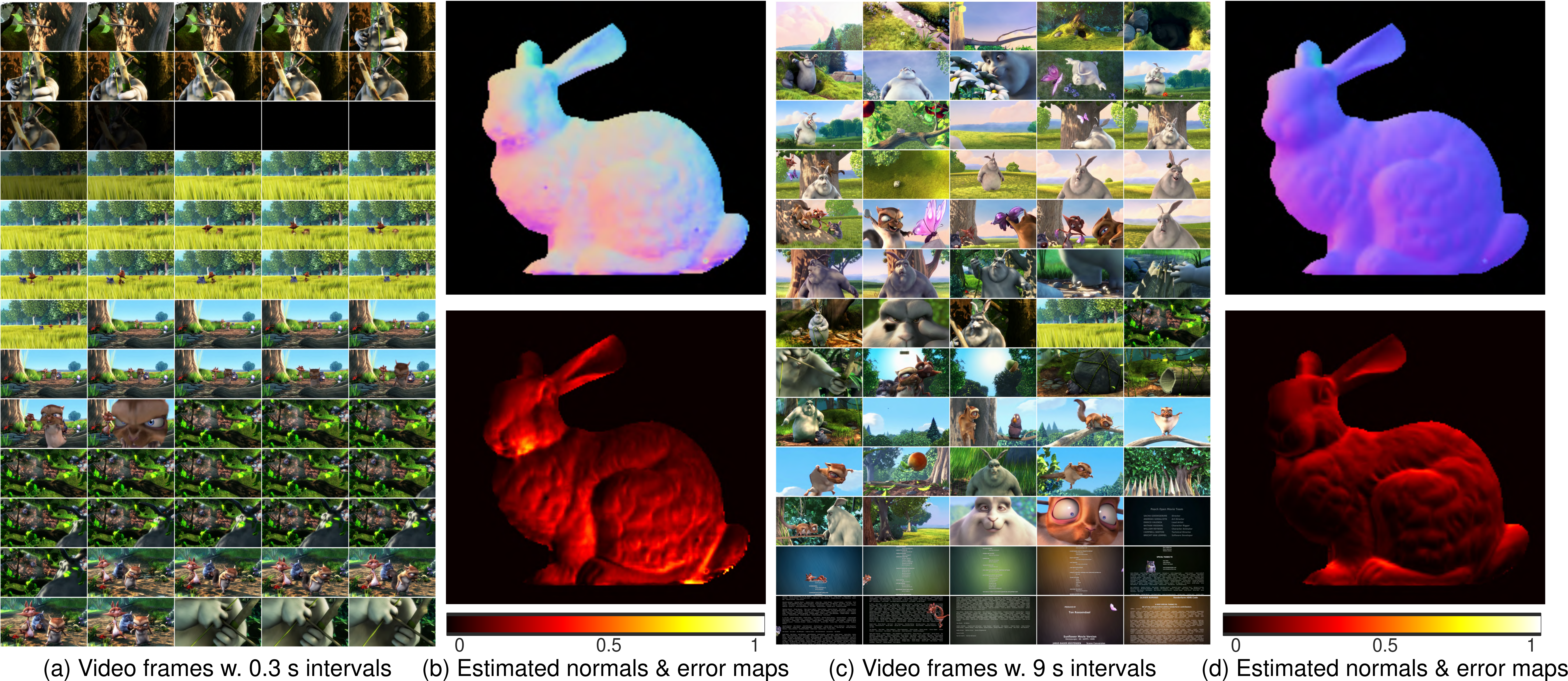}
\caption{\textbf{Reconstruction with video frames.} (a) Video frame set with 0.3 second intervals shows (b) estimated normals (top) and error maps (bottom). (c) Video frame set with 9 second intervals shows (d) estimated normals (top) and error maps (bottom).}
\label{fig:video_frame}
\end{figure*}

We reconstruct normals using two different 65-frame set sampled from the \textit{Big Buck Bunny} video where the intervals within frame are 0.3/9 seconds respectively.
Figure~\ref{fig:video_frame} shows frames, reconstructed normals, and error maps.
The inter-frame standard deviations of each frame set are 0.2094/0.2658 respectively.
We use 65 frames, which can be stored in our GPU memory limit.
The reconstruction errors are 0.1893/0.1140 for the two videos, showing the dependency of video contents on reconstruction accuracy.
In contrast, using learned patterns with DDPS achieves the reconstruction error of 0.0476 with only two patterns (Tri-random, Table~\ref{tab:num_illum}).
Dynamic photometric stereo with two learned patterns would be an interesting future work.

\subsection{Generalizability of Learned Patterns}
{
We conduct cross validation for demonstrating the generalizability of DDPS.
Table~\ref{tab:cross_validation} shows that DDPS achieves consistently low reconstruction errors and similar characteristics of learned patterns for the cross-validation test. }

\begin{table}[h]
\centering
\resizebox{0.55\columnwidth}{!}{
    \begin{tabular}{c|c}
    \hline
    (Mean/std. dev.) of recon. error & (Mean/std. dev.) of learned-pattern intensity \\ \hline 
     (0.0457/0.0054)   &(0.4371/0.0842)   \\ 
    \hline
    \end{tabular}
}
\caption{{\textbf{Statistics of reconstruction error and learned patterns.} 5-fold cross validation shows consistent reconstruction error and learned-pattern intensity with low std. dev..}}
\label{tab:cross_validation}
\end{table}

\subsection{Scene Geometry Assumption}
{
Due to inaccessibility of accurate geometry of the inference scene, we assume that the surface points of objects lie on a plane located 50\,cm away from the camera position along the z-axis.
This assumption is critical for normal estimation in conventional methods as it interrupt utilizing ground-truth lighting vectors.
To demonstrate the robustness of DDPS under such assumption, we conduct a comparison experiment between patterns learned using ground-truth lighting vectors from ground-truth depth and patterns learned using proposed method.
In the former case, it shows reconstruction error of 0.0467 with using GT depth.
In comparison, DDPS achieves reconstruction error 0.0475 without using GT depth.
These results implicate the robustness of DDPS in learning patterns that can compensate deviations in scene geometry assumptions.}

\subsection{Display Size}
We test DDPS on a simulated 32'' display by sampling 5$\times$10 central superpixels of the original 55’’ display.
Normal reconstruction learned and tested on the 32'' display gives the reconstruction error of 0.0659, when using mono-gradient initialization.
Even though this error is larger than that of using the original 55’’ display (0.0443), the learned patterns enables outperforming the best reconstruction accuracy of 0.0805 on a 55’’ display using heuristic patterns, group OLAT. 

\subsection{Generalizability to Arbitrary In-the-wild Shapes}
\begin{figure*}[h]
\centering
\includegraphics[width=0.65\textwidth]{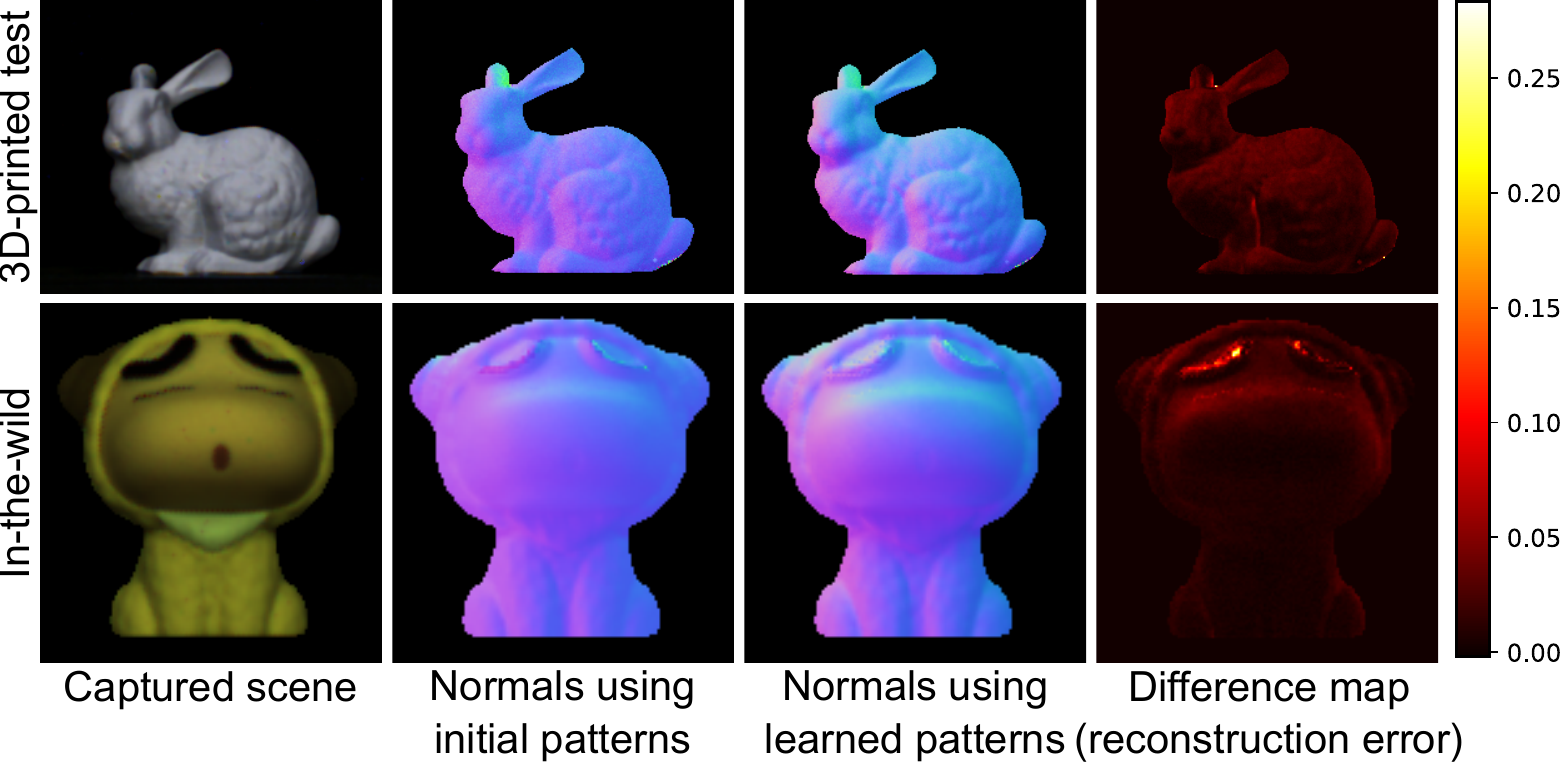}
\caption{\textbf{DDPS for test/in-the-wild objects.}}
\label{fig:inthewild}
\end{figure*}

{Figure~\ref{fig:inthewild} shows that learned patterns improves normal reconstruction for in-the-wild objects.}

\subsection{Comparison with Learning-based Photometric Stereo}
\begin{figure*}[h]
\centering
\includegraphics[width=\textwidth]{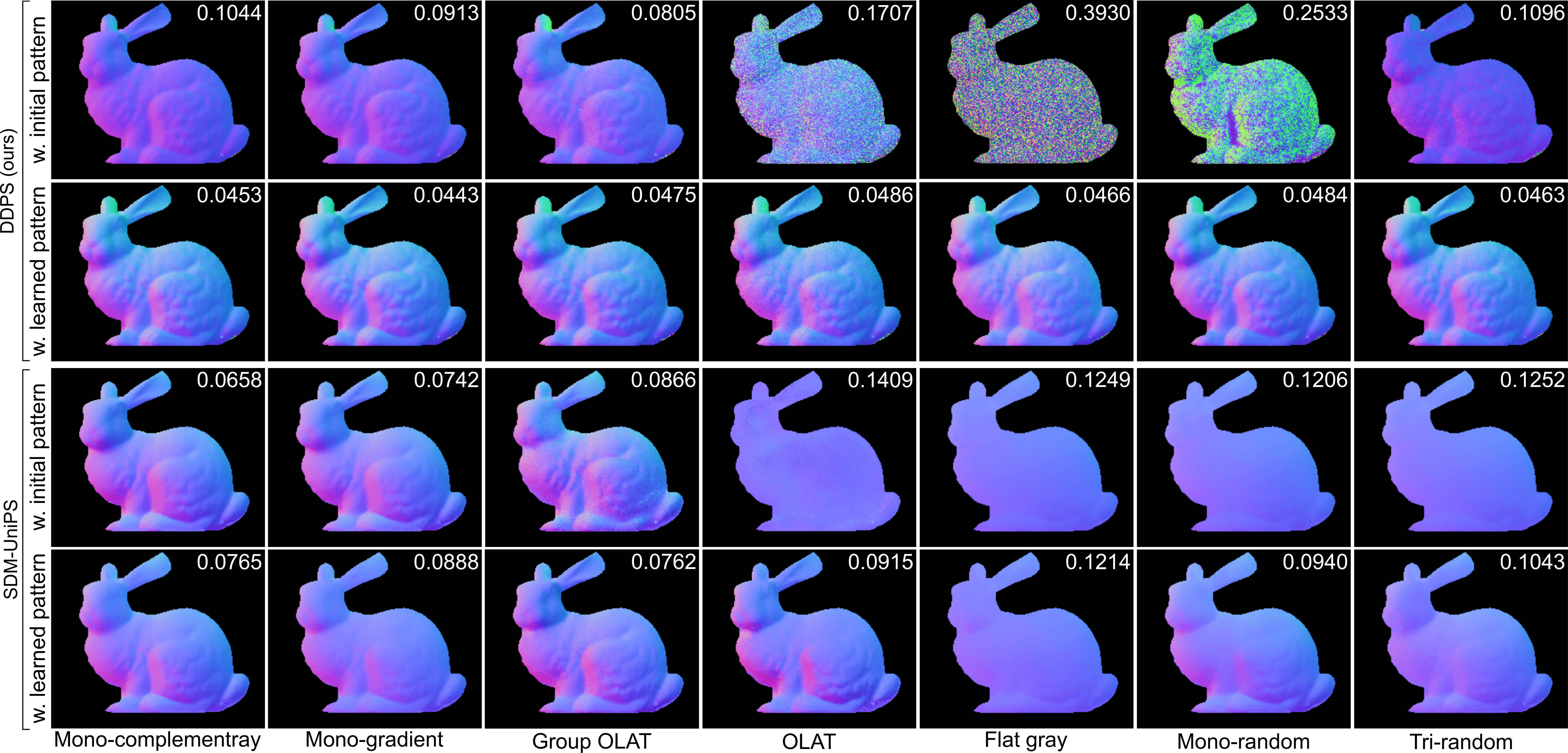}
\caption{\textbf{Comparison with SDM-UniPS.} The reconstruction error is indicated in the upper right corner of the each reconstructed normal. The uncalibrated learning-based methods often fails on leveraging lighting context.}
\label{fig:sdm_unips}
\end{figure*}

We compare the reconstruction results with DDPS to state-of-the-art normal reconstruction method, SDM-UniPS~\cite{ikehata2023scalable} that leverages neural networks.
SDM-UniPS uses mask-free inputs and supports unknown and arbitrary lighting conditions.
Figure~\ref{fig:sdm_unips} show the reconstruction results of SDM-UniPS on our test dataset.
The uncalibrated learning-based methods are often fragile with complex lighting contexts or out-of-distribution objects, and also cannot fully leverage benefits of carefully-designed illumination.
In contrast, the physically-valid reconstruction methods enables DDPS robust on aforementioned cases.

\section{Additional Results}
We show additional results of DDPS in Figures~\ref{fig:results_supp} and \ref{fig:results_supp2}, including captured images, their respective illumination patterns, surface normals, and diffuse albedo.
We also provide a failure example due to strong highlights. 

\subsection{Results with Different Learned Patterns}
\begin{figure*}[h]
\centering
\includegraphics[width=0.75\textwidth]{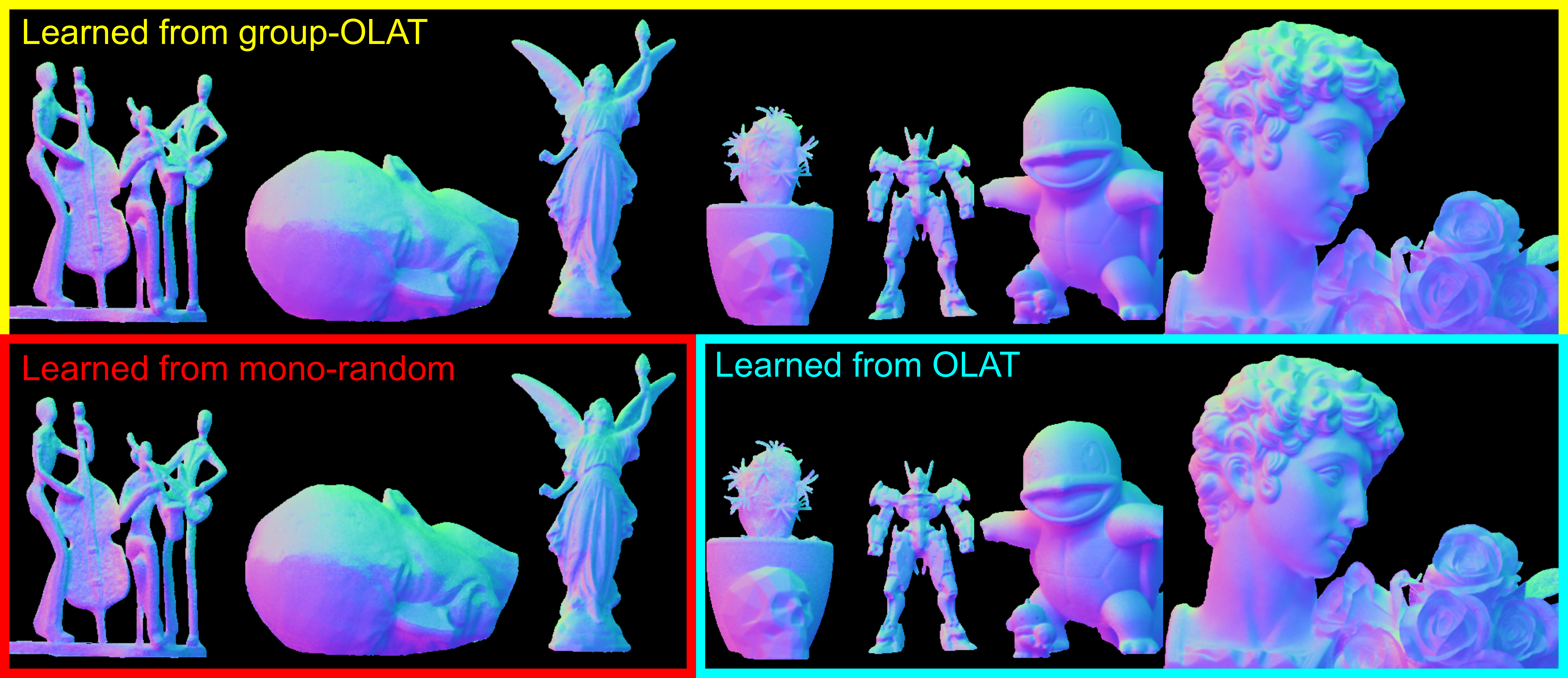}
\caption{\textbf{Results with different learned patterns (top vs. bottom).} DDPS shows similar qualitative reconstruction results with different learned patterns.}
\label{fig:diff_patterns}
\end{figure*}

{Figure~\ref{fig:diff_patterns} shows that reconstruction results with different learned patterns are generally similar. Section~\ref{global_illumination} further shows that severe inter-reflection in a concave bowl makes notable difference between learned-pattern results.}

\subsection{Diffuse Albedo}
\begin{figure*}[t]
\centering
\includegraphics[width=\textwidth]{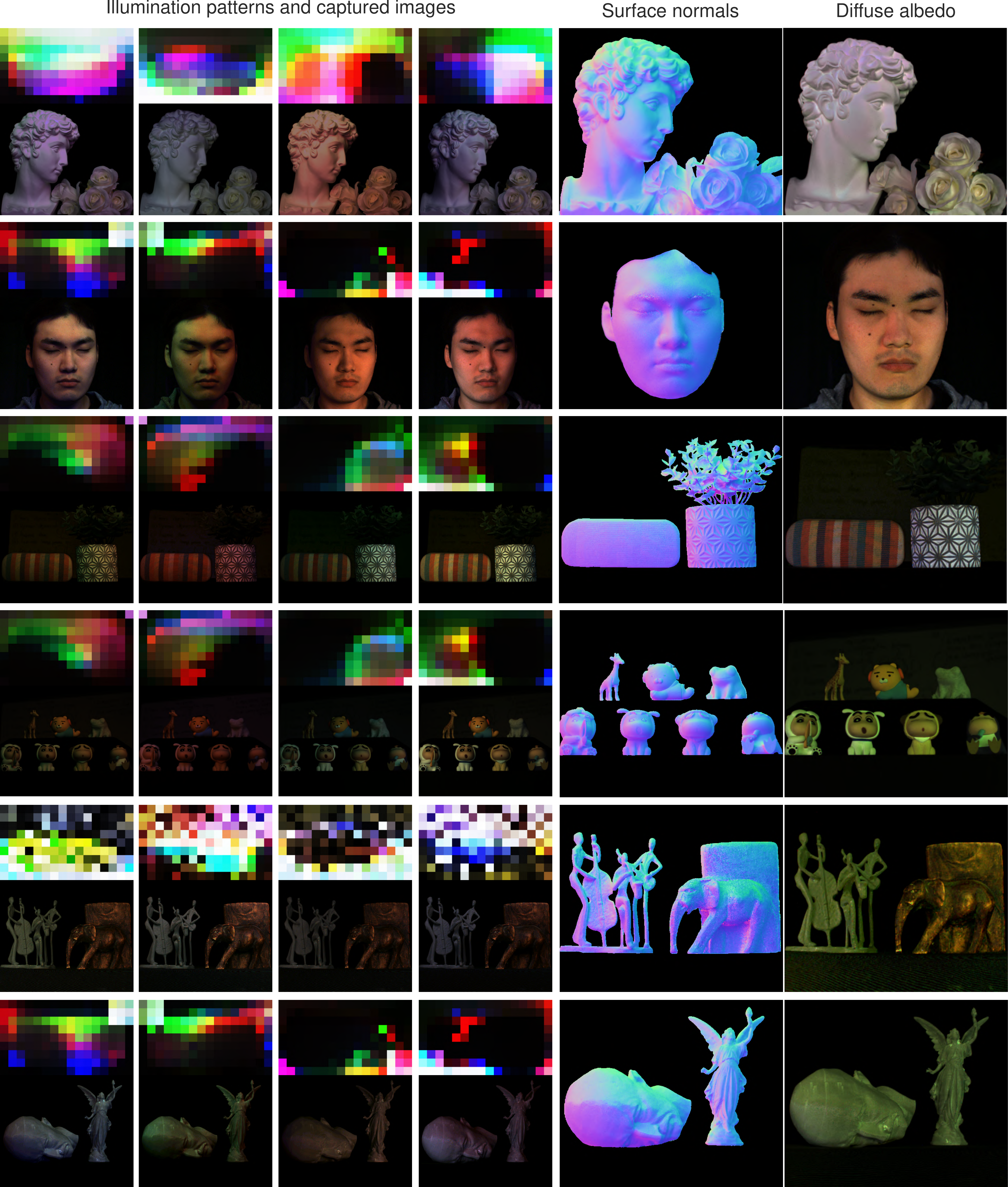}
\caption{\textbf{Additional results of DDPS.}}
\label{fig:results_supp}
\end{figure*}

Figure~\ref{fig:results_supp} and \ref{fig:results_supp2} shows the reconstructed surface normals and diffuse albedo of various objects including a human face from four input images captured using the learned patterns.

\subsection{Robustness against Ambient Illumination}
\begin{figure*}[t]
\centering
\includegraphics[width=\textwidth]{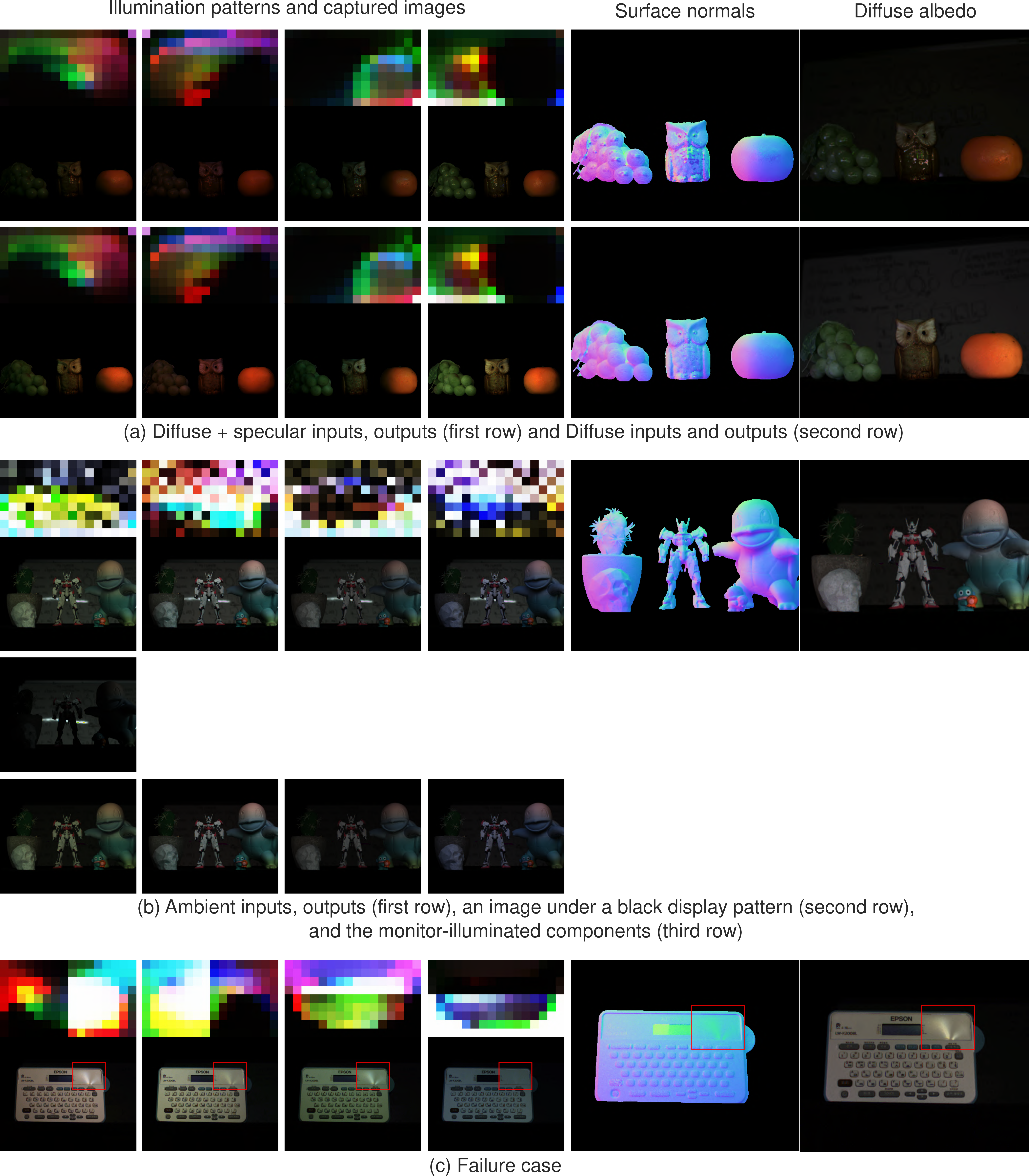}
\caption{\textbf{Additional results of DDPS.}}
\label{fig:results_supp2}
\end{figure*}

We experimentally demonstrate testing our learned patterns while ambient light is present. To this end, we capture an additional image under a black display pattern to capture the contribution only from ambient light. We then subtract this ambient-only image from the images taken under the learned display patterns with ambient light. This enables isolating the display-illuminated components only. We then use photometric-stereo reconstruction for obtaining surface normals. To handle the limited dynamic range of the display and the camera, we use HDR imaging for obtaining high-quality normal reconstruction. Figure\ref{fig:results_supp2} shows the reconstructed surface normals.

\clearpage

\clearpage
{\small
\bibliographystyle{ieee_fullname}
\bibliography{supple}
}

\end{CJK}